\documentclass[letterpaper]{article}
\usepackage{proceed2e}
\usepackage[margin=1in]{geometry}

\usepackage{booktabs} 

\usepackage{algorithm}
\usepackage{algorithmic}
\usepackage{amsmath}
\usepackage{amsthm}
\usepackage{mathrsfs}
\usepackage{amssymb}
\usepackage{graphicx}
\usepackage{subfigure}
\usepackage{bm}
\usepackage{amsopn}
\usepackage{threeparttable}
\usepackage{multirow}
\usepackage{booktabs}
\usepackage{url}
\usepackage{cleveref}

\title{Automatic Ensemble Learning for Online Influence Maximization}

\author{
  Xiaojin Zhang\textsuperscript{\rm 1}\\
  \textsuperscript{\rm 1}The Chinese University of Hong Kong,
  xjzhang@cse.cuhk.edu.hk}

\begin{document}

\maketitle

\begin{abstract}

We consider the problem of selecting a seed set to maximize the expected number of influenced nodes in the social network, referred to as the \textit{influence maximization} (IM) problem. We assume that the topology of the social network is prescribed while the influence probabilities among edges are unknown. In order to learn the influence probabilities and simultaneously maximize the influence spread, we consider the tradeoff between exploiting the current estimation of the influence probabilities to ensure certain influence spread and exploring more nodes to learn better about the influence probabilities. The exploitation-exploration trade-off is the core issue in the multi-armed bandit (MAB) problem. If we regard the influence spread as the reward, then the IM problem could be reduced to the combinatorial multi-armed bandits. At each round, the learner selects a limited number of seed nodes in the social network, then the influence spreads over the network according to the real influence probabilities. The learner could observe the activation status of the edge if and only if its start node is influenced, which is referred to as the edge-level semi-bandit feedback. Two classical bandit algorithms including Thompson Sampling and Epsilon Greedy are used to solve this combinatorial problem. To ensure the robustness of these two algorithms, we use an automatic ensemble learning strategy, which combines the exploration strategy with exploitation strategy. The ensemble algorithm is self-adaptive regarding that the probability of each algorithm could be adjusted based on the historical performance of the algorithm. Experimental evaluation illustrates the effectiveness of the automatically adjusted hybridization of exploration algorithm with exploitation algorithm.
\end{abstract}
%
%

\section{Introduction}

Social network is very effective in propagating information through word-of-mouth, which provides a powerful avenue for the marketers to expand the influence of their products. With certain amount of budget, the marketers are willing to provide some influential users with free products, in hope that they could influence more users in the social network to purchase the products. The final goal is to select a set of nodes maximizing the expected number of influenced users. This kind of problem is referred to as the \textit{influence maximization} (IM) problem. The selected nodes are called the \textit{seed set}. 

The influence propagation could be modeled as a stochastic cascade, two commonly used cascade models are \textit{Independent Cascade} (IC) and \textit{Linear Threshold} (LT) \cite{kempe2003maximizing}. For the IC model, the influence propagation is independent of the historical process. Initially, the seed nodes are activated. Each activated node at the current step has one and only one chance to activate its neighborhoods at the next step, and the activated node could not turn back to the state of being inactivated. This process continues until no further activation is available. For the LT model, the node is activated if the sum of weights of its activated neighborhoods exceeds a threshold. The IM problem under these models are proved to be NP-hard. As a result, a great number of work proposed heuristics \cite{chen2010scalable, chen2009efficient, jung2012irie, kim2013scalable} and approximation algorithms \cite{leskovec2007cost, goyal2011celf++}.

Classical influence maximization researches assume that the influence probabilities between nodes are prescribed \cite{vaswani2015influence}.  However, it is usually impractical to obtain the influence probability in many application scenarios. The algorithm needs to learn the influence probability in order to facilitate the selection of the most influential nodes. On the one hand, the past selections and feedbacks should be exploited to guarantee certain amounts of influence spread. On the other hand, those nodes which have not be sampled sufficiently are encouraged to be explored in order to raise the learner's awareness about the global network. The exploration-exploitation tradeoff fits into the framework of multi-armed bandits and motivates a line of research on IM semi-bandits \cite{lei2015online, chen2016combinatorial, vaswani2015influence,wen2016influence,carpentier2016revealing,vaswani2017diffusion}.  

To make the algorithm scalable to large social networks, \cite{wen2016influence, vaswani2017diffusion} made linear generalization assumption on the influence probability under the contextual combinatorial bandit framework, and proposed the LinUCB based algorithm for edge-level feedback and pairwise feedback separately. For the edge-level feedback, a limited number of seed sets are selected at each round, and the activation status of each edge whose start node is influenced could be observed. With the feedback and the contextual information, the estimation of the influence probabilities could be updated, and the updated influence probabilities are then used to select the seed nodes at the next round.

Thompson Sampling is one of the earliest heuristics for the MAB problem, which randomizes action according to their uncertainty, i.e. it choses the best arm based on the parameter sampled from the posterior distribution. \cite{scott2010modern, chapelle2011empirical} demonstrated the good empirical performance of Thompson Sampling algorithm, which greatly raised the researcher's interest in Thompson Sampling. The first theoretical results about Thompson Sampling was proposed in \cite{agrawal2012analysis}, then a series of work including \cite{kaufmann2012thompson} and \cite{korda2013thompson} have been proposed, these works show that Thompson Sampling achieves optimal performance in the frequentist setting \cite{abeille2017linear}. Besides, \cite{russo2016information} mentioned that Thompson Sampling is very effective regarding to the linear bandits \cite{tossou2017thompson}.

We firstly propose a Thompson Sampling based algorithm for the IM problem in the contextual combinatorial bandit framework, under the assumption that the influence probability is linear with the contextual information. The linear realizability assumption is made to ensure certain efficiency of the algorithm. However, it might be hard for the linear function to explain the relationship between the influence probability and the contextual information. In view of this situation, we hope to devise an efficient algorithm without the assumption of any particular functional form. As a result, we further propose a novel algorithm without the linear realizability assumption, the algorithm is easy to implement and is both efficient in terms of the running time and effective regarding to the influence spread.

For the exploitation strategy, it takes good advantage of the past observations, while it is easy for this kind of algorithm to be trapped in the locally optimal solution. The exploration strategy has a higher probability to find the global optimal solution, meanwhile this kind of algorithm might face higher risk. The complementary nature of these two categories of algorithms motivates the hybridization of the exploitation methods with the exploration approaches. We introduce an ensemble learning framework that combines the exploitation with the exploration approaches automatically. Under this framework, the algorithm used in each round is sampled according to the probability distribution, and the probability associated with each algorithm depends on the historical influence spreads gained by selecting this algorithm. Experimental results on real datasets illustrate that our method significantly outperforms the state-of-the-art algorithms in terms of the influence spread. Besides, our algorithm tends to be more robust to various datasets since the preferences towards the strategies are updated promptly according to the performance of the algorithm on the dataset. 

\section{Related Work}

\cite{domingos2001mining, richardson2002mining} firstly provided the probabilistic methods for influence maximization problem. \cite{kempe2003maximizing} then proved the IM problem to be NP-hard, and they showed that this problem could be approximated with a constant approximation guarantee of $1-1/e$. Since then, various improved algorithms have been proposed, including heuristics with high efficiency (Degree Discount \cite{chen2009efficient}, IRIE \cite{kim2013scalable}), and approximation algorithms with theoretical guarantee (CELF \cite{leskovec2007cost}, CELF++ \cite{goyal2011celf++}). Currently, TIM and TIM+ \cite{tang2014influence} and IMM \cite{tang2015influence} are the most widely used algorithms.

Considering the influence probabilities are unknown in many application scenarios, a line of work on IM semi-bandits was proposed to learn the influence probability and at the same time maximizing the influence spread \cite{lei2015online, chen2016combinatorial, vaswani2015influence,wen2016influence,carpentier2016revealing, vaswani2017diffusion}. The feedbacks of these bandit algorithms could be divided into three categories according to the feedback \cite{valko2016bandits}: 1) full-bandit, where the learner could only observe the number of influenced nodes; 2) edge-level feedback, in which the diffusion status of each edge is observed; 3) node-level feedback, in which the learning agent could only observe the activation status of each node, but have no idea about the specific diffusion process. \cite{lei2015online, wen2016influence, vaswani2017diffusion}  focus on the edge-level feedback, while \cite{vaswani2015influence} proposed an algorithm with regard to the node-level feedback.

The work including \cite{fang2014networked}, \cite{wen2016influence}, \cite{saritacc2016online} and \cite{vaswani2017diffusion} exploited the contextual information to deal with the influence maximization problem. \cite{fang2014networked} assume that the reward of an arm is obtained from the selected user and the neighborhoods. The arm is selected based on the contextual information and the network relationship, but they did not provide any formal definition about the relationship. \cite{wen2016influence} proposed a LinUCB-based bandit algorithm \textit{IMLinUCB} with regard to each edge, and \cite{vaswani2017diffusion} proposed a LinUCB-based bandit algorithm \textit{DILinUCB} with regard to each pair of nodes. They made the linear assumption in order to ensure that the algorithm is efficient for large-scale IM semi-bandits.

\cite{lei2015online} considered the IM problem in the edge-level semi-bandit feedback setting, and showed that the $\epsilon$-greedy algorithm and the Confidence-Bound (CB) strategies have good empirical performance. \cite{vaswani2015influence} provided theoretical analysis for both the edge-level feedback and the node-level semi-bandit feedback. \cite{chen2016combinatorial} proposed a UCB-based algorithm called CUCB for the edge-level semi-bandit feedback, they show that the reward function of the influence maximization is monotonous and bounded-smooth, and thus their algorithm CUCB for the combinatorial semi-bandits could be applied to the IM problem, while the regret bound might be exponentially large for influence maximization bandits. 

\section{The Models}

\subsection{Influence Maximization}

The social network could be represented by a directed graph $G=(V,E)$, where $V$ represents the individuals, and $E$ represents the relationships among the individuals. The graph is directed considering that the influences among people are asymmetric. The goal of the influence maximization problem is to select a set of nodes that maximizes the expected number of influenced nodes, satisfying that the number of seed nodes not exceeding a specified value $K$ due to some considerations including the budget constraint \cite{valko2016bandits}. Denote by $\mathbf w^{*}$ the real influence probabilities among edges, $\mathbf y$ corresponds to the activation status of the edges, which is sampled according to the influence probability, i.e. $\mathbf y(e) \sim $Bern($\mathbf w^{*}(e)$). The influence maximization problem could be formally described as 
\begin{align}
\arg\max_{S\subseteq V} f(S, \mathbf w^{*}) =\sum_{v\in\uppercase{V}} \mathbb E[ \mathbb I(S,\mathbf y, v)],
\end{align}
where $f(S,\mathbf w^{*})$ the expected number of nodes influenced by the seed set $S$, $\mathbb I(S,\mathbf y,v)$ is an indicator function that takes on a value of 1 if node $v$ is activated under the seed set $S$ and the particular activation status vector $\mathbf y$, and 0 otherwise. Note that the expectation is taken over the randomness of the sampled activation status $\mathbf y(e)$ for $e\in E$, and the possible randomness of the IM algorithm. 

\subsection{Contextual Combinatorial Bandits}
The problem of contextual combinatorial bandits could be formulated as follows.  At each round $t$, the learner is presented with a set of arms $\mathcal{A} = \{1,2,\dots,K\}$, each arm $e\in \mathcal{A}$ is associated with a feature vector $\mathbf x_{t}(e)$, and the learner could select a set of arms satisfying the number of the selected arms not exceeding a specified number. The leaner decides which arm set to choose at round $t$ based on the historical information: the actions and the corresponding rewards at rounds $\tau<t$, and the feature vectors of each arm at rounds $\tau\le t$. Denote these historical information up to round $t$ as 
\begin{align}
&\mathcal{H}_{t} = \{a_{\tau}, w_{\tau}(a_{\tau}), \mathbf x_{\tau}(e), \mathbf x_{t}(e)| e\in [K], \tau\in[t-1]\}. 
\end{align}

Under the linear reward assumption, the expected reward of each arm is a linear combination of the contextual information. Denote by $w_{t}^{*}(e) = \mathbb{E}[w_{t}(e)|\mathcal{H}_{t}] $, then the expected reward of arm $e$ at round $t$ could be formally represented as 
\begin{align}
&w_{t}^{*}(e) = \bm{\theta}_{*}^{\mathrm T} \mathbf x_{t}(e). 
\end{align}

The goal of the learner is to choose the arm set with the highest cumulative reward, or equivalently, to minimize the cumulative regret. Denote by $w_{t}^{*}$ the real linear coefficient, the cumulative regret is defined as 
\begin{align}
&R(T) = \sum_{t=1}^{T} [f(A_{t}^{*}, \mathbf{w}_{t}^{*}) - f(A_{t}, \mathbf{w}_{t}^{*})],
\end{align}
where $A_{t}$ is the arm set selected at round $t$ based on the estimated weight vector $\mathbf{w}_{t}^{*}$, and $A_{t}^{*} = \arg\underset{A\subset\mathcal{A}}{\max}\, f(A, w_{t}^{*})$ is the arm set with the highest expected reward at round $t$, referred to as the optimal arm set.

\section{Online Influence Maximization}

In many practical scenarios, the topology of the social network is available while the exact influence probabilities among users are unknown. The problem of maximizing the influence when the influence probabilities are unavailable is referred to as online influence maximization (OIM) \cite{lei2015online}. To accumulate information about the influence probabilities in the social network, the influence is propagated for multiple rounds with various selections of the seed set, and the feedback information is collected to refine the learning about the influence probabilities. Under the edge-level feedback, the activation status of the edges whose start nodes are influenced could be observed. The learning agent needs to exploit the past observations and feedbacks to guarantee certain influence spread, and explore more nodes to have a better understanding about the whole social network. Taking the contextual information associated with each node into consideration, then the online IM problem fits into the framework of contextual combinatorial bandits. Each edge in the social network could be regarded as an arm. At round $t$, a set of nodes are selected. Each selected node could influence their direct neighborhoods, and the activated node could influence their direct neighborhoods at the next step, this proceeds until no node will be further activated. For each influenced node $s$, the learner could observe the activation status $y(s,v)$, $\forall  v\in\uppercase{V}$. The activation status of each edge (arm) could be regarded as the reward of this edge. The total number of activated nodes by the source node set $S$ is regarded as the reward of $S$. Considering that the IM problem is NP-hard, the regret is the gap between the scaled number of activated nodes by the optimal set $S^{*}$ and the number of activated nodes by the selected seed set $S_{t}$. The regret for the IM semi-bandits could be represented as

\begin{align}
R^{\eta}(T) = \eta Tf(S^{*}, w^{*}) - \sum_{t=1}^{T} f(S_{t}, w^{*}),
\end{align}
where $w^{*}$ is the true influence probability vector and is unknown to the learning agent.

\textbf{Algorithm}
\textbf{\ref{Framework: The framework of online influence maximization}} illustrates the basic framework of the online influence maximization. The algorithm runs for a total of $T$ rounds. $X$ represents the feature vector associated with the edges, $A_t$ is the arm set selected at round $t$, $P_t^{'}$ represents the estimated influence probabilities of the edges at round $t$, $F_{t}$ represents the statistics analyzed from the feedback information, and \textit{ORACLE}  is the offline IM algorithm. The basic framework of OIM problem contains the following three main parts: 
\begin{itemize}
\item \textbf{Influence Estimation(line \ref{alg_oim_estimate}):} The learner estimates the influence probabilities based on the contextual information and the feedback information; 
\item \textbf{Seed Selection (line \ref{alg_oim_SeedSelection}):} The seed set is selected by the offline oracle algorithm with the estimated influence probabilities;
\item \textbf{Information Processing (line \ref{alg_oim_InfSpread}):} The feedback information is collected, and the collected information including the activated edges is processed and embodied in the statistics. For the contextual setting, the information also includes the contextual information for each node.
\end{itemize}

It was shown by \cite{goyal2011data} that the learning of the influence probabilities from the propagation information is of critical importance in maximizing the influence. In the following sections, we propose several algorithms for influence estimation, and analyze the performance of these algorithms.
\begin{algorithm}
\caption{The framework of online influence maximization}
\label{Framework: The framework of online influence maximization}
\begin{algorithmic}[1]
\STATE{\textbf{Input: }$ORACLE, G, K, X$}
\FOR{$t\leftarrow1,2,...,T$}
\STATE{$\hat P_{t} \leftarrow ESTIMATE(X, A_{1:t-1})$}\label{alg_oim_estimate}
\STATE{$S_{t}\leftarrow ORACLE (G, \hat P_{t}, K)$}\label{alg_oim_SeedSelection}
\STATE{$F_{t} \leftarrow FEEDBACK (G, S_{t})$}\label{alg_oim_InfSpread}
\ENDFOR
\end{algorithmic}
\end{algorithm}

\section{CCMAB Framework for OIM}
Denote by $A_{t}$ the set of nodes that are influenced by the seed set $S$ at round $t$. Under the edge-level feedback, at each round $t$ the learner could observe the activation status of the edge $e = (u,v)$ whose start node $u\in A_{t}$, i.e. node $u$ is influenced at round $t$, denote as $E_{t}^{'}$ (line \ref{alg2_edge_set1}). $N_{t, e}$ represents the total number of times edge $e = (u,v)$ has been activated up to round $t$, and $T_{t, e}$ denotes the total number of times the activation status of edge $e = (u,v)$ has been observed until round $t$, i.e. the total number of times node $u$ has been influenced until round $t$. The online influence maximization algorithm under the contextual combinatorial bandits with linear payoff assumption is displayed in \textbf{Algorithm} \textbf{\ref{alg:The OIM algorithm under the CCMAB Framework}}. 

At each round $t$, the influence probabilities are estimated based on the contextual information and the feedback information (line \ref{alg2_estimate}). With the estimated influence probability, a seed set $S$ is selected by the oracle algorithm (line \ref{alg2_Seed_Selection}). Then the influence propagates across the network according to the independent cascade model. With the feedback information and the contextual information, $V_{t}$ and $Y_{t}$ could be updated accordingly (line \ref{alg2_V_Update} and line \ref{alg2_Y_Update}).

Under the assumption that the influence probability is linear with the contextual information, the linear coefficient vector could be estimated based on the influence probability at each round. In fact, the feedback information we could get is the activation status of the edges whose start nodes are influenced (line \ref{alg2_feedback}). Then we estimate the influence probability using Thompson Sampling (line \ref{alg2_TS}). The activation status of each edge follows the Bernoulli distribution, i.e. edge $e$ is activated with the influence probability $p(e)$. Assume that the prior distribution of the influence probability of each edge follows the Beta distribution, i.e.  $p(e) \sim$ Beta ($\alpha$, $\beta$), then the posterior distribution of $p(e)$ at round $t$ follows from Beta ($\alpha+N_{t, e}$, $\beta+T_{t, e}-N_{t, e}$).

\begin{algorithm}
\caption{The OIM algorithm under the CCMAB Framework}
\label{alg:The OIM algorithm under the CCMAB Framework}
\begin{algorithmic}[1]
\STATE{\textbf{Input: }$\alpha,\beta, ORACLE, G, K, X$}
\STATE{\textbf{Initialization: }$\hat{\theta}_{0}\leftarrow 0_{d\times 1},V_{0}\leftarrow \lambda I_{d\times d},Y_{0} \leftarrow 0_{d\times 1}$}
\FOR{$t\leftarrow1,2,...,T$}
\STATE{$\hat P_{t} = ESTIMATE(V_{t-1}, Y_{t-1}, X)$}\label{alg2_estimate}
\STATE{$S_{t}\leftarrow$ ORACLE ($G$, $C$, $\hat P_{t}$)}\label{alg2_Seed_Selection}
\STATE{//INFORMATION PROCESSING}
\STATE{$V_{t}\leftarrow V_{t-1}+\sum_{e\in E}x_{t}(e)x_{t}(e)^\mathrm{T}$}\label{alg2_V_Update}
\STATE{$E_{t}^{'} \leftarrow \{(u,v): (u,v)\in E, u\in A_{t}\}$}\label{alg2_edge_set1}
\FOR{$e\in E_{t}^{'}$}
\STATE{$T_{t, e} \leftarrow T_{t-1, e} + 1$}
\STATE{get feedback $n_{t, e}$} \label{alg2_feedback}
\STATE{$N_{t, e} \leftarrow N_{t-1, e} + n_{t, e}$}
\STATE{sample $\widetilde p_{t}(e)$ from Beta ($\alpha$+$N_{t, e}$, $\beta+T_{t, e}-N_{t,e}$)}\label{alg2_TS} 
\ENDFOR
\STATE{$Y_{t}\leftarrow Y_{t-1}+\sum_{e\in E_{t}^{'}}x_{t}(e)\widetilde p_{t}(e)$}\label{alg2_Y_Update}
\ENDFOR
\end{algorithmic}
\end{algorithm}

We consider two strategies to estimate the influence probability of each edge under the contextual combinatorial framework with the linear realizability assumption: the Upper Confidence Bound (UCB) strategy and the Thompson Sampling (TS) strategy. We will illustrate these strategies in detail.

\subsection{Upper Confidence Bound Strategy}
With the information gained during the online process, the learner could learn the relationship between the contextual information and the reward. Denote the edge set in the social network as $E$. In the contextual combinatorial setting, the learner is provided with the contextual information $x_{t}(a^{t}_{e})$ associated with each edge $e$ for $e\in E$ at round $t$. A set of nodes could be selected as the seed set at each round, then the activation status of the edges whose start nodes are influenced could be observed according to the edge-level feedback. At round $t$, denote the total number of edges whose activation status could be observed is $K_{t}$, then the learner could observe the reward $w_{t}(a^{t}_{i})$ of each edge $a^{t}_{i}$ for $i\in [K_{t}]$. Under the assumption of linear payoff, the upper bound of the expected reward could be estimated using the ridge regression.  Specifically, denote $\mathbf W_{t} = [\mathbf W_{t-1}; w_{t}(a^{t}_{1}); \dots ; w_{t}(a^{t}_{K_{t}})], \mathbf X_{t} = [\mathbf X_{t-1}; x_{t}(a^{t}_{1}); \dots ; x_{t}(a^{t}_{K_{t}})]$, where $K_{t}$ is the number of selected arms at round $t$, and $\mathbf V_{t} = \mathbf X_{t}^{\mathrm{T}}\mathbf X_{t} + \lambda I_{d}$, then the upper confidence bound of each edge $e$ for $e\in E$ at round $t$ could be represented as
\begin{align}
U_{t}(e) = \hat\theta(t)^{\mathrm{T}} x_{t}(e) + \beta \sqrt{x_{t}(e)^{\mathrm{T}}\mathbf V_{t}^{-1} x_{t}(e)},
\end{align}
where $\bm\hat\theta_{t} = \mathbf V_{t}^{-1}\mathbf X_{t}^{\mathrm{T}}\mathbf W_{t}$ and $\beta >0$.
The ridge regression estimation could be regarded as the mean estimation $\hat w_{t}(a^{t}_{i})$ added with the width estimation $\hat\sigma_{t}(a^{t}_{i})$. The width estimation could be interpreted as the Mahalanobis distance from the center of mass of $\mathbf X_{t}$, which has the property that the distance is smaller when $x_{t}(a^{t}_{i})$ is closer to the center of mass of $\mathbf X_{t}$. Specifically, when the noise follows the standard normal distribution, the width corresponds to the standard deviation of $\hat w_{t}(a^{t}_{i})$. In fact, $\beta$ could be a value updated along the rounds, \cite{abbasi2011improved} proposed a UCB based algorithm with the confidence radius $\beta_{t} = \sqrt{\lambda} + \sqrt{\log (|\mathbf V_{t-1}|/(\lambda^{d}\delta^{2}))}$ at round $t$, and the regret bound of which is $O(d\log(T)\sqrt{T}+\sqrt{dT\log(T/\delta)})$, where $d$ is the dimension of the feature vector.

\subsection{Thompson Sampling Strategy}

In the contextual bandit setting such as the news recommendation problem, Thompson Sampling algorithm tends to have better performance than the other approaches such as UCB \cite{agrawal2013thompson}, which motivates us to use the Thompson Sampling algorithm to estimate the influence probabilities based on the contextual information and the feedback information. The paper \cite{agrawal2013thompson} proposed a Thompson Sampling based algorithm for linear contextual bandits with linear payoff assumption, the high probability regret bound of which is $\widetilde O(\min\{d^{3/2}\sqrt{T}, d\sqrt{Tlog(N)}\})$. At each round $t$, the algorithm \textit{LinThompson} samples $\widetilde\theta_{t}$ from a multi-variate Gaussian distribution $\mathcal{N}(\hat\theta_{t}, v^{2}\mathbf V_{t}^{-1})$, and the arm with the maximal $x_{t}(e)^{\mathrm{T}}\widetilde\theta_{t}$ is selected. For convenience of description, we call this algorithm the \textit{LinThompson} algorithm. 

To make the \textit{LinThompson} algorithm applicable to the OIM problem, we extend this algorithm to the contextual combinatorial setting. In terms of the linear payoffs assumption, the expected reward for each edge $e$ is a linear combination of the contextual information. Specifically, $w_{t}^{*}(e) = \theta_{*} ^{\mathrm{T}}x_{t}(e)$, where $\theta_{*}$ is the linear coefficient vector and is unknown to the learner. Suppose that the likelihood of the reward $w_{t}(e)$ of arm $e$ at time $t$ is $\mathcal{N}(\theta_{*}^{\mathrm{T}}x_{t}(e), v^{2})$, where $v = R\sqrt{9d\log(t/\delta)}$, $d$ is the dimension of the feature vector. If we set the prior for $\theta_{*}$ at time $t$ as $\mathcal{N}(\hat\theta_{t}, v^{2}\mathbf V_{t}^{-1})$, then the posterior distribution for $\theta_{*}$ at time $t+1$ is given by $\mathcal{N}(\hat\theta_{t+1}, v^{2}\mathbf V_{t+1}^{-1})$. The definition of $\mathbf V_{t}$ and $\hat\theta_{t}$ is illustrated in Eq.\eqref{eq_V} and Eq. \eqref{eq_hat_theta}, where $a^{\tau}_{k}$ represents the $k^{th}$ edge whose activation status could be observed at round $\tau$, and $K_{\tau}$ is the total number of edges whose activation status could be observed at round $\tau$.

\begin{align}
&V_{t} = I_{d} + \sum_{\tau=1}^{t-1} \sum_{k=1}^{K_{\tau}}x_{\tau}(a^{\tau}_{k}) x_{\tau}(a^{\tau}_{k})^{\mathrm{T}}\label{eq_V}\\
&\hat\theta_{t} = V_{t}^{-1}(\sum_{\tau=1}^{t-1} \sum_{k=1}^{K_{\tau}} x_{\tau}(a^{\tau}_{k}) w_{t} (a^{\tau}_{k})).\label{eq_hat_theta}
\end{align}

\subsection{The LinThompsonUCB Algorithm}

We observe that if the influence probability is directly estimated using the extended \textit{LinThompson} strategy, then the selected seed set often falls into a locally optimal solution, which implies that the exploration tendency embedded in the pure TS strategy is not sufficient. To make the algorithm more suitable for the OIM problem, we merged the exploration part of the UCB strategy with the TS strategy, the pseudo-code of the proposed \textit{LinThompsonUCB} algorithm is illustrated in \textbf{Algorithm} \textbf{\ref{alg: LinThompsonUCB Algorithm}}. The influence probability of edge $e$ is initially estimated using $x_{t}(e)^{\mathrm{T}}\widetilde\theta_{t}$, where $\widetilde\theta_{t}$ is sampled according to the Thompson Sampling strategy. The confidence interval is further merged to the estimation to make this algorithm more suitable for the online influence maximization problem, where $\beta_{t}$ is an effective confidence radius proposed by \cite{abbasi2011improved}.

\begin{algorithm}
\caption{LinThompsonUCB algorithm}
\label{alg: LinThompsonUCB Algorithm}
\begin{algorithmic}[1]
\STATE{\textbf{Input: } $\mathbf V_{t-1}, \mathbf W_{t-1}, \mathbf X$}
\STATE{\textbf{Output: $\hat p_{t,e}$ for $e\in E$} }
\STATE{//ESTIMATE}
\STATE{$\hat{\theta}_{t-1} \leftarrow \mathbf V_{t-1}^{-1}\mathbf W_{t-1}$}
\STATE{$\beta_{t-1} \leftarrow \sqrt{\lambda} + \sqrt{\log (|\mathbf V_{t-1}|/(\lambda^{d}\delta^{2}))}$}
\STATE{Sample $\widetilde\theta_{t}$ from $\mathcal{N} (\hat\theta_{t-1}, v^{2}\mathbf V_{t-1}^{-1})$}
\FOR{$e\in E$}
\STATE{$\hat p_{t, e} \leftarrow  x_{t}(e)^ \mathrm{T} \widetilde\theta_{t} + \beta_{t-1} \Vert x_{t}(e) \Vert_{\mathbf V_{t-1}^{-1}}$}
\ENDFOR
\end{algorithmic}
\end{algorithm}

\section{CMAB Framework for OIM}
For some algorithms, it is commonly seen that the algorithm performs excellent in one dataset, but might have poor performance in another dataset. Sometimes the algorithms may not perform well just due to the fact that the parameters of the algorithms have not been set suitably, while it is sometimes hard or require lots of energy to find the parameters that well-matched with the specific dataset or problem. In view of this situation, we ensemble the pure exploration and pure exploitation strategies, and adjust the preference towards these algorithms automatically based on the performance of the algorithms. It was shown by \cite{seldin2012pac} that the practical performance of \textit{EXP3.P} is significantly inferior to that of \textit{EXP3} in the stochastic setting, which motivates the use of \textit{EXP3} algorithm to select the strategies.

\begin{algorithm}
\caption{EXP3 algorithm}
\label{alg: Weight Adjust Algorithm}
\begin{algorithmic}[1]
\STATE{\textbf{Initialization: } probability distribution $\psi_{i} = 1/N$, weights $w_{i} = 1$ for strategy $i = 1,2,\dots,K$ } 
\FOR{$t\leftarrow1,2,...,T$}
\STATE{$g_{t-1} \leftarrow \sum_{v\in\uppercase{V}} \mathbb I(S_{t-1},\mathbf y_{t-1}, v)/|V|$}
\FOR{$i\leftarrow1,2,...,N$}
\STATE{$w_{i}\leftarrow w_{i}\times \exp(\gamma\times (g_{t-1} \times \mathbb I[i= I_{t-1}])/\psi_{i})$}
\ENDFOR
\STATE{$W\leftarrow \sum_{k=1}^{N} w_{k}$}
\FOR{$i\leftarrow1,2,...,N$}
\STATE{$\psi_{i}\leftarrow (1-\gamma)\times w_{i}/W + \gamma\times 1/n$}
\ENDFOR
\ENDFOR
\end{algorithmic}
\end{algorithm}

\begin{algorithm}
\caption{UpdateExp3}
\label{alg: Weight Adjust Algorithm}
\begin{algorithmic}[1]
\STATE{\textbf{Input:}  influence spread $C$, selected strategy $M$, weight vector $\bm{w}$, probability distribution $\bm{\psi}$}
\STATE{\textbf{Output:} weight vector $\bm{w_{'}}$, probability distribution $\bm{\psi_{'}}$} 
\STATE{$g \leftarrow C/|V|$}
\FOR{$i\leftarrow1,2,...,N$}
\STATE{$w'_{i}\leftarrow w_{i}\times \exp(\gamma\times (g \times \mathbb I[i = M])/\psi_{i})$}
\ENDFOR
\STATE{$W'\leftarrow \sum_{k=1}^{N} w'_{k}$}
\FOR{$i\leftarrow1,2,...,N$}
\STATE{$\psi'_{i}\leftarrow (1-\gamma)\times w'_{i}/W' + \gamma\times 1/N$}
\ENDFOR
\end{algorithmic}
\end{algorithm}

Assume there are $N$ base strategies. Denote by \bm{$w$} = $(w_{1}, w_{2},\dots,w_{N})$ the weight vector, \bm{$\psi$} = $(\psi_{1}, \psi_{2},\dots,\psi_{N})$ the probability distribution over the strategies, satisfying that $\psi_{i}\ge 0, i = 1,2,\dots, N$, and $\Vert \bm{\psi} \Vert_{1}= 1$. The \textit{EXP3} algorithm starts with the probability distribution  $\psi_{i} = 1/N$, and the weight vector $w_{i} = 1$, $i = 1,2, \dots, N$. If the selected strategy leads to a high influence spread at current round, then the \textit{EXP3} algorithm tends to assign this strategy with higher weight, and improve the probability of using this strategy at the next round.

In fact, the thought of determining which strategy to use based on the probability distribution is also embedded in the design of \textit{Epsilon-Greedy} algorithm. There are a total of two strategies to be selected, one is the exploit strategy, and the other is the explore strategy. The probability distribution for these two strategies is $(\epsilon, 1-\epsilon)$. One of the main distinctions of these two algorithms lies in that the probability distribution for the \textit{Epsilon-Greedy} algorithm has to be set manually and is prescribed, while for the Ensemble Learning algorithm, the probability distribution could be adjusted automatically based on the observed feedback.

  \begin{figure*} 
\centering 
\subfigure[Subgraph of Facebook]{
  \label{fig1:distinct:c} 
 \includegraphics[width = 0.3\linewidth]{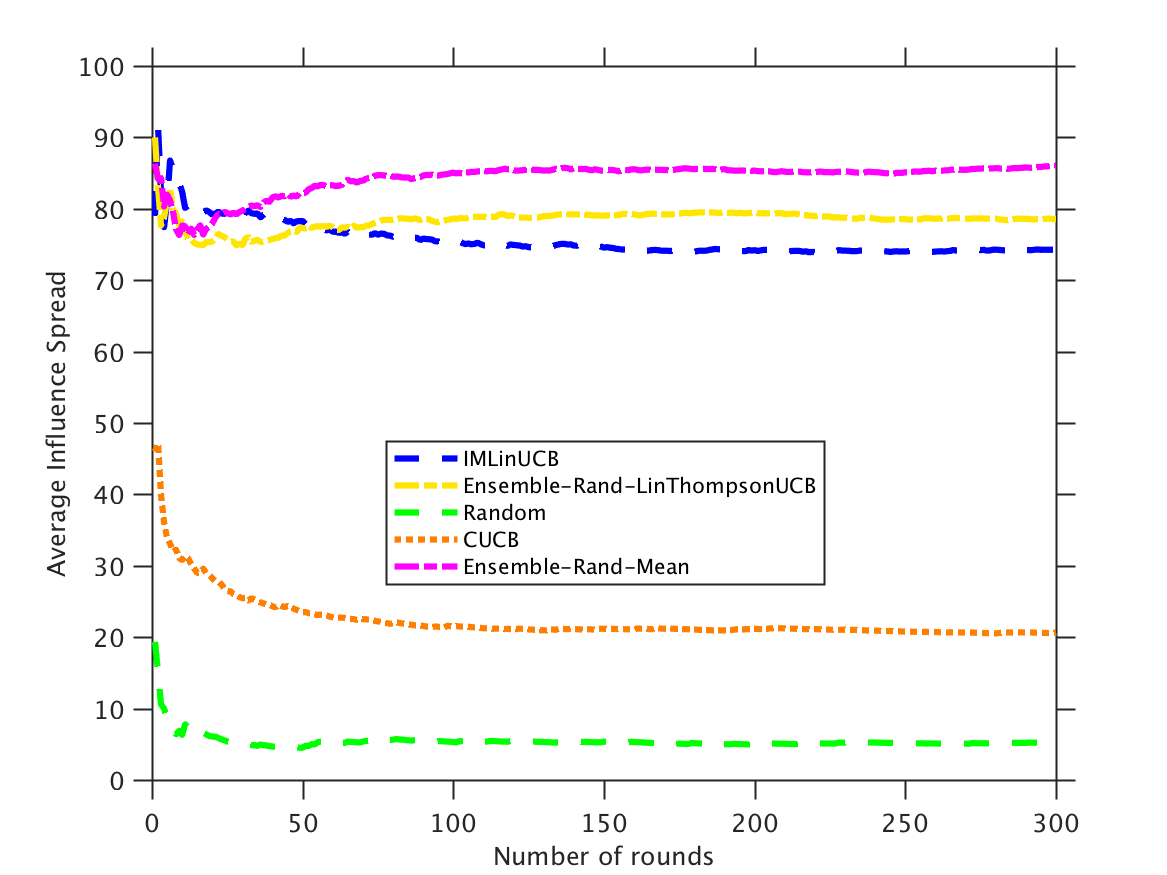}} 
\subfigure[Facebook]{
 \label{fig1:distinct:a} 
 \includegraphics[width = 0.3\linewidth]{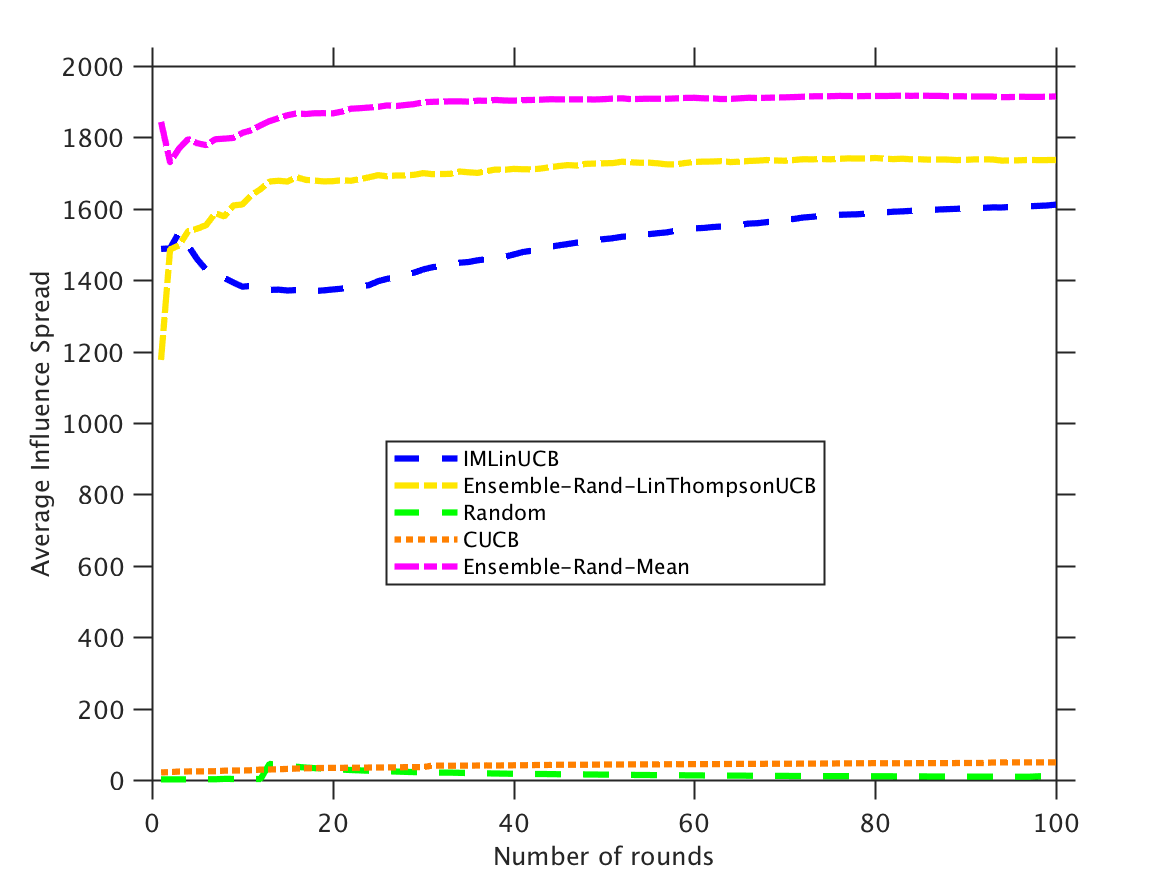}} 
 \subfigure[NetHEPT]{
  \label{fig1:distinct:b} 
 \includegraphics[width = 0.3\linewidth]{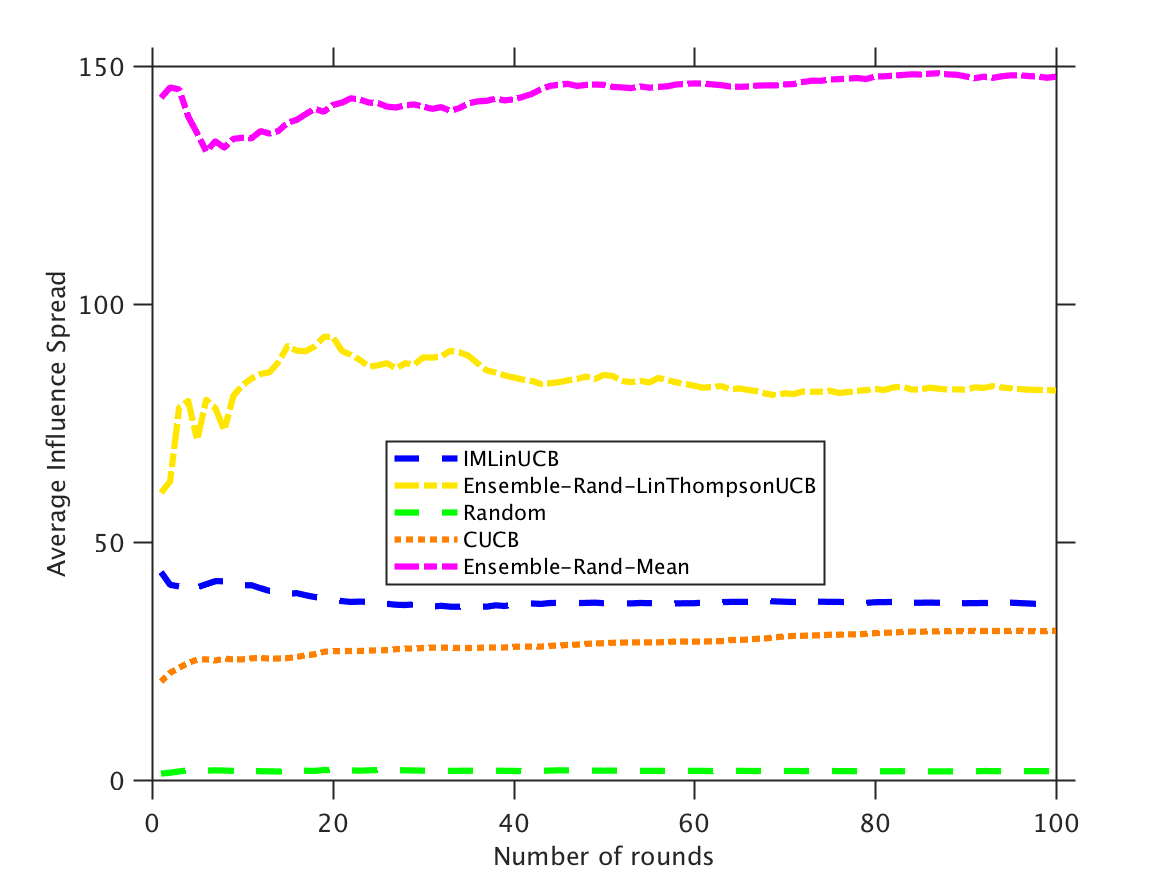}} 
 \caption{Average influence spread for distinct algorithms}
 \label{fig_aver_inf_distinct} 
 \end{figure*} 

   \begin{figure*} 
\centering 
\subfigure[Subgraph of Facebook]{
  \label{fig1:distinct:c} 
 \includegraphics[width = 0.3\linewidth]{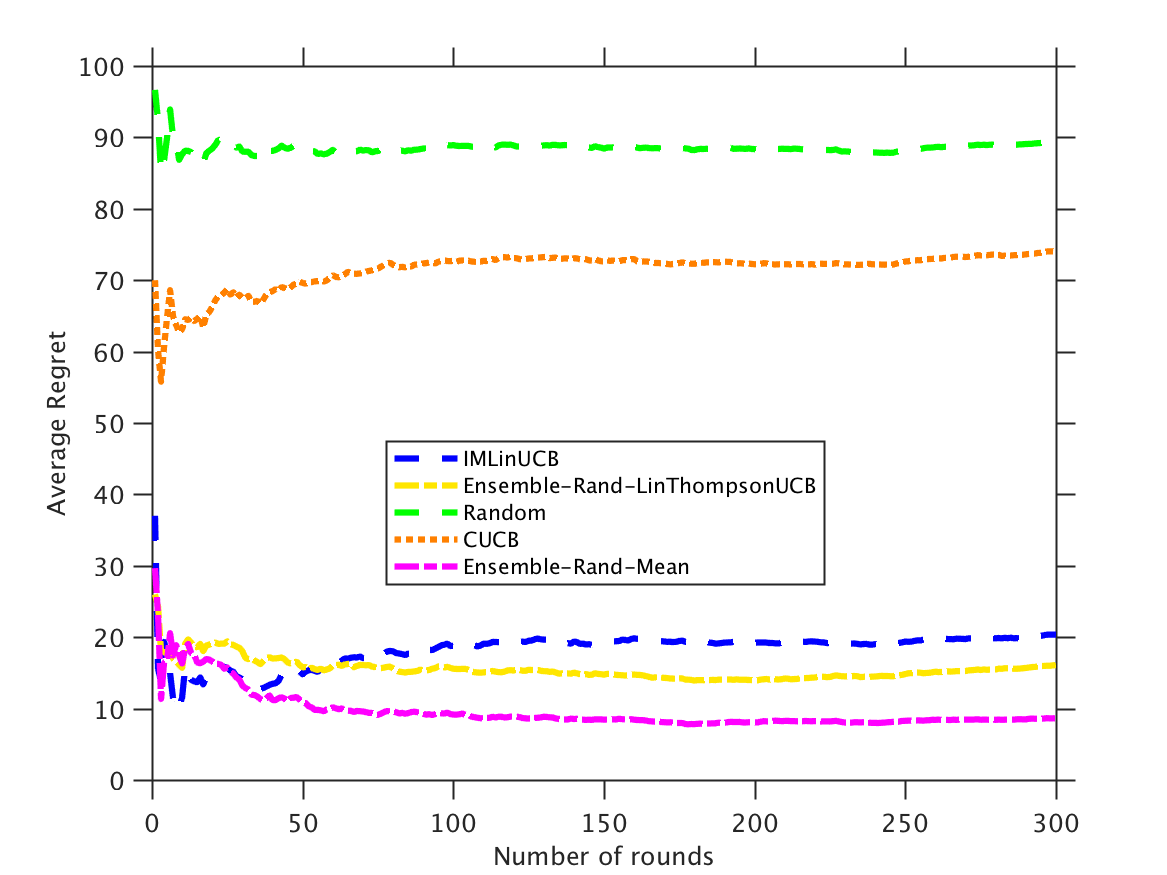}} 
\subfigure[Facebook]{
 \label{fig1:distinct:a} 
 \includegraphics[width = 0.3\linewidth]{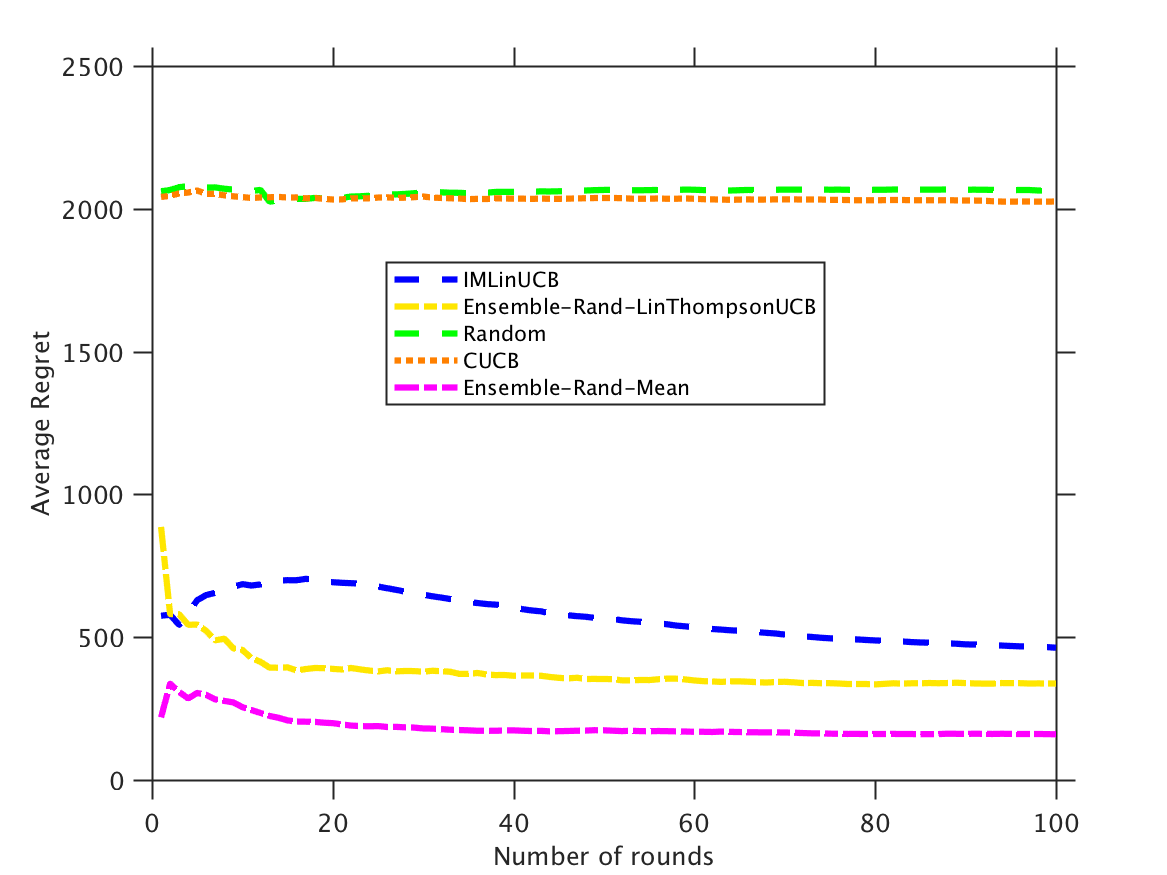}} 
 \subfigure[NetHEPT]{
  \label{fig1:distinct:b} 
 \includegraphics[width = 0.3\linewidth]{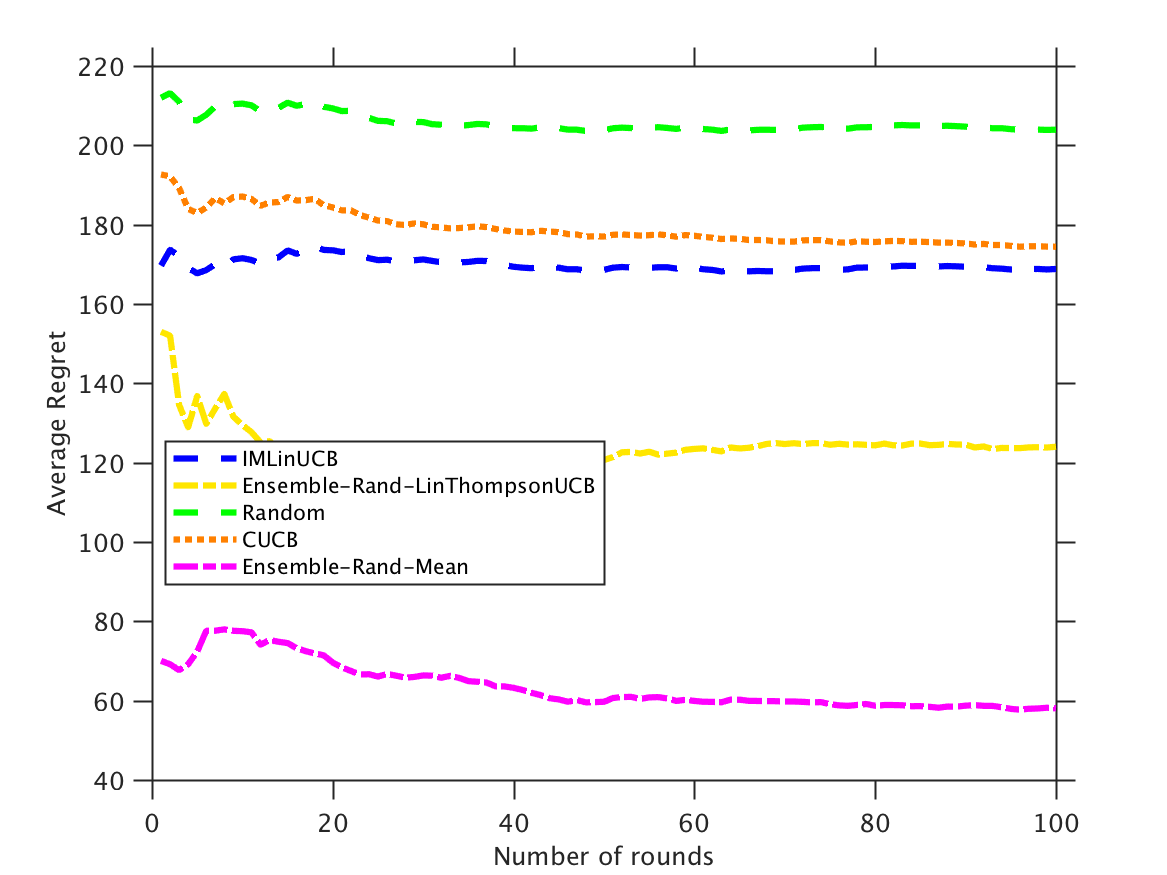}} 
 \caption{Average regret for distinct algorithms}
 \label{fig_aver_regret_distinct} 
 \end{figure*} 

\begin{algorithm}
\caption{Ensemble-Rand-Mean Algorithm}
\label{alg: rand-mean algorithm}
\begin{algorithmic}[1]
\STATE{\textbf{Initialization: } probability distribution $\psi_{i} = 1/N$, weights $w_{i} = 1$ for strategy $i = 1,2$ } 
\FOR{$t\leftarrow1,2,...,T$}
\STATE{Sample the strategy $M_{t}$ randomly according to the probability distribution $\bm{\psi_{t}}$}
\IF{$M_{t}$ is $1$}
\FOR{$e\in E$}
\STATE{$\hat{p}_{t, e} \leftarrow X_{t, e}/T_{t, e}$}\label{alg_Ensemble_mean}
\ENDFOR
\ELSIF{$M_{t} $ is $2$}
\FOR{$e\in E$}
\STATE{$\hat{p}_{t, e} \leftarrow rand(0,1)$}\label{alg_Ensemble_rand}
\ENDFOR
\ENDIF
\STATE{$S_{t}\leftarrow$ ORACLE($G$, $\hat P_{t}$, $K$)}\label{alg_Ensemble_oracle}
\STATE{$C_{t}, A_{t}\leftarrow$ CALSPREAD$(G, P^{*}, S_{t})$}\label{alg_Ensemble_calspread}
\STATE{$[\bm{w}_{t+1},\bm{\psi}_{t+1}] \leftarrow$ UpdateExp3$(C_{t}, M_{t}, \bm{w}_{t}, \bm{\psi}_{t})$}\label{alg_Ensemble_weightupdate}
\STATE{$E^{'} = \{(u,v): (u,v)\in E, u\in A_{t}\}$}\label{alg_Ensemble_EQ}
\FOR{$e\in E^{'}$}
\STATE{$T_{t+1, e} \leftarrow T_{t, e} + 1$}
\STATE{get feedback $x_{t, e}$} \label{alg_Ensemble_feedback}
\STATE{$X_{t+1, e} \leftarrow X_{t, e} + x_{t, e}$}
\ENDFOR
\ENDFOR
\end{algorithmic}
\end{algorithm}

Denote $M_{t}$ as the label of the selected method at round $t$, $C_{t}$ as the influence spread at round $t$. The pseudo-code of \textit{Ensemble-Rand-Mean} is illustrated in \textbf{Algorithm} \textbf{\ref{alg: rand-mean algorithm}}. A total of two strategies are provided for selection. Initially, the probability of selecting both strategies are identical. At round $t$, the learner samples a strategy according to the probability distribution \bm{$\psi_{t}$}, and estimate the influence probabilities based on the selected strategy. If the selected strategy is strategy 1, then we exploit the empirical mean to estimate the influence probability (line \ref{alg_Ensemble_mean}). If the selected strategy is strategy 2, then we explore more nodes by estimating the influence probability using the value randomly sampled from $\mathcal U(0,0.01)$ (line \ref{alg_Ensemble_rand}).  Denote by $\hat P_{t}$ the estimated influence probabilities for the edges at round $t$, $P^{*}$ the real influence probabilities associated with the edges. With the estimated influence probabilities, the seed set $S_{t}$ is selected by the offline oracle algorithm (line \ref{alg_Ensemble_oracle}), then the influence spread is calculated using the real influence probability (line \ref{alg_Ensemble_calspread}), and the influence spread observed at this round is utilized to update the weight of each strategy (line \ref{alg_Ensemble_weightupdate}).

\section{Experiments}

In this section, we conduct experiments to show the performance of distinct algorithms for solving the OIM problem under the independent cascade (IC) model, and analyze the strength of the automatic ensemble learning strategy. The experiments illustrate the significant improvement of our algorithm over the compared state-of-art algorithms. Besides, the \textit{Ensemble-Rand-Mean} algorithm is efficient with respect to both the running time and space consuming, and this algorithm is robust regarding to various datasets.

\subsection{Experimental Settings}

The experiment results are averaged across three independent simulations. Note that in practical, each round of experiment requires the company to offer $K$ influencers with free product, and collect the edge-level feedback information, which is rather costly. Thus, the total number of running rounds will not be set as very large such as the order of thousands in the experiments. Next we will illustrate the implementation details about the experiments including the construction of the feature vector, the generation of the influence probability, the simulation of the influence propagation, and the selected oracle algorithm.

  \begin{figure*} 
\centering 
 \subfigure[Subgraph of Facebook]{
  \label{randthomp} 
 \includegraphics[width = 0.41\linewidth]{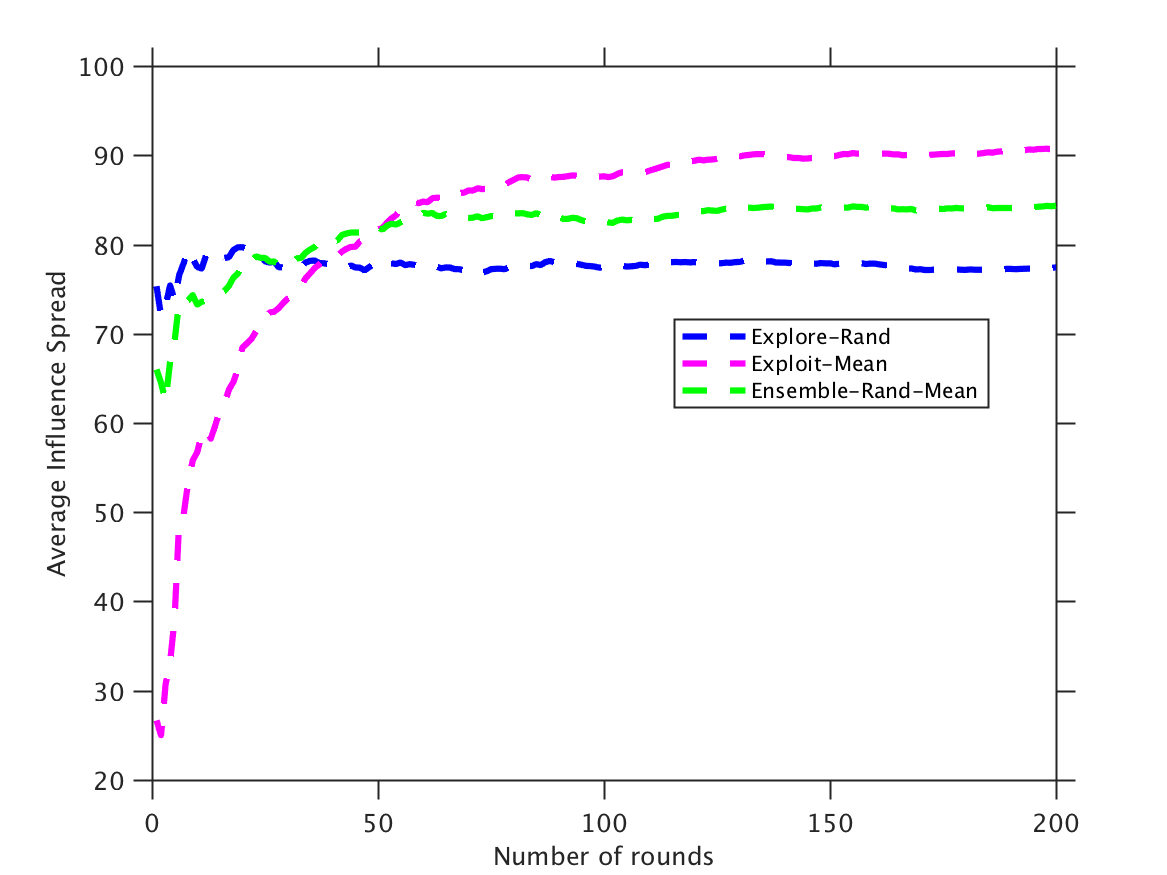}} %
\subfigure[Facebook]{
 \label{randmean} 
 \includegraphics[width = 0.41\linewidth]{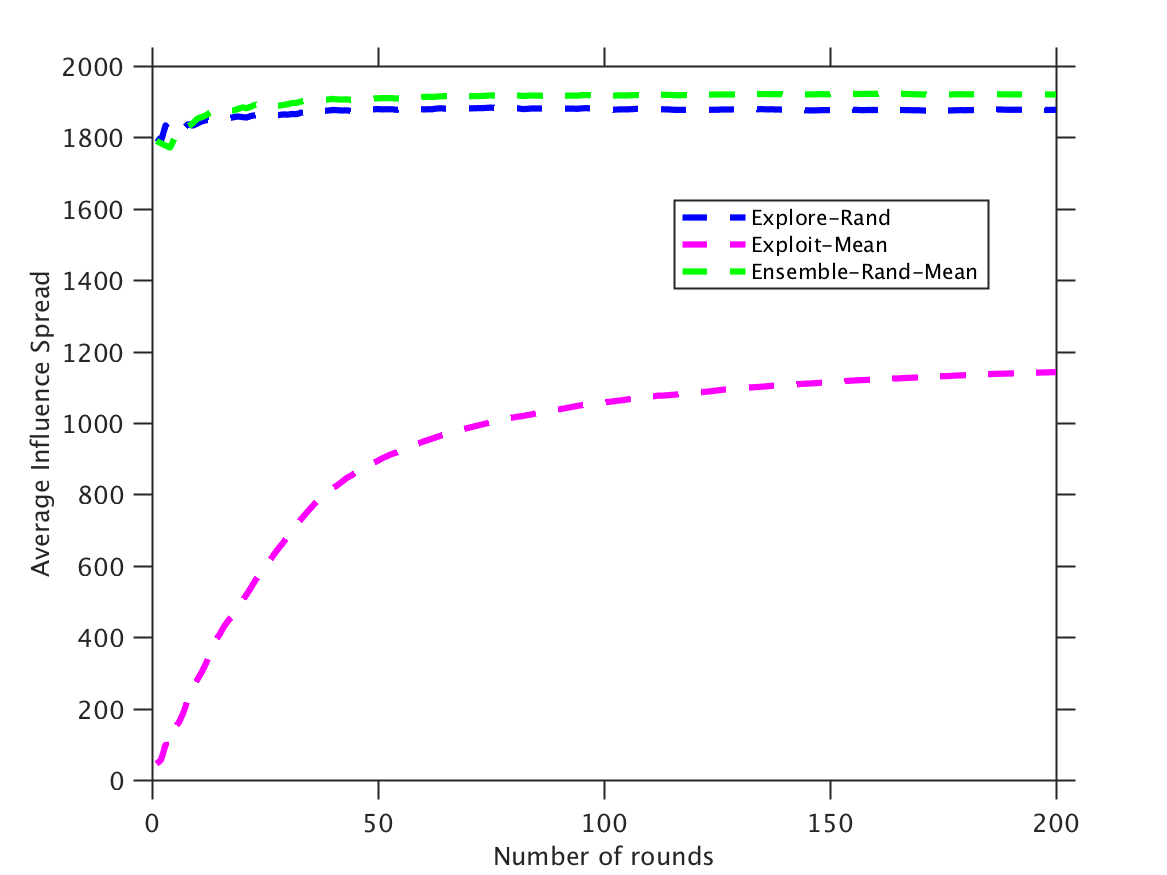}} 
 \caption{\textit{Ensemble-Rand-Mean} V.S. \textit{Explore-Rand} and \textit{Exploit-Mean}}
 \label{figRandMean_Rand_Mean} 
 \end{figure*}

\textbf{Datasets:}  In the experiment, we used three datasets to evaluate the performance of the algorithms, separately the Facebook dataset, the NetHEPT dataset, and the subgraph of the facebook dataset. The Facebook dataset is extracted from survey participants using the Facebook app, it contains 4039 nodes and 88,234 edges. The NetHEPT dataset is a real world academic collaboration network, which is extracted from the "High Energy Physics - Theory" section of arXiv \footnote{http://www.arXiv.org}. This dataset contains 15,233 nodes and 58,891 unique edges, and each node represents an author, each edge represents the coauthorship relation. In order to embody the power of the algorithm from distinct aspects, we further extracted a subgraph of the Facebook dataset, which contains 347 nodes and 5,038 edges.

\textbf{Feature vector:} The feature vector of node $u$ is constructed using the laplacian approach, denote as $f(u)$. For the edge $(u,v)$, the feature vector $f(u,v)$ of the edge is the pointwise product of its two endpoints, i.e. $f(u,v) = f(u)*f(v)$.

\textbf{Influence probability:}  The influence probability associated with edge $(u,v)$ in the graph is set as $1/degree(v)$, which has been commonly used in many papers including \cite{goyal2011data, kempe2003maximizing, tang2014influence, leskovec2007cost}.

\textbf{Influence propagation:} The activation status of each edge follows the Bernoulli distribution. Suppose the influence probability of edge e is p(e), then edge e is activated with probability p(e), and inactivated with probability 1-p(e). In term of the implementation, we sample a number randomly for each edge, then the activation status of each edge is determined by the random number and the true influence probability. Specifically, if the random number is greater than the influence probability, then this edge is regarded as activated; otherwise, this edge is regarded as non-activated. Given a seed set, then the spread on the graph could be regarded as the diffusion over the sampled deterministic graph.

\textbf{Oracle algorithm:} The oracle algorithm provides an ($\alpha$-$\beta$) - approximation solution. Specifically, given the topology of the graph and the influence probability, the oracle algorithm could output an $\alpha$- approximation to the optimal value with probability $\beta$. The oracle algorithm used for the experiments is the state-of-the-art algorithm \textbf{TIM}, which is proposed by \cite{tang2014influence}.  This algorithm returns a $(1-1/e-\epsilon)$ - approximate solution with at least $1-n^{-l}$ probability.

\subsection{Performance of the Proposed Algorithms}

In the first experiment, we compare the performance of our proposed \textit{Rand-LinThompson} and \textit{Rand-Mean} algorithms with the following algorithms, which includes both the contextual dependent algorithms and the algorithms that do not use the contextual information.
\begin{itemize}
\item\textbf{Random}: This algorithm selects a set of $K$ nodes randomly from the $n$ nodes in the graph.

\item\textbf{CUCB}: CUCB is an algorithm that uses the UCB strategy in the combinatorial multi-armed bandit (CMAB) setting, which is proposed by \cite{chen2016combinatorial}.
\item\textbf{IMLinUCB}: IMLinUCB is also a UCB-based algorithm, which utilizes the contextual information and makes the linear generalization assumption, proposed by \cite{wen2016influence}.
\end{itemize} 
The performance of distinct algorithms in terms of the average influence spread and the average regret on various datasets are shown in \textbf{Figure}  \textbf{\ref{fig_aver_inf_distinct}} and \textbf{Figure}  \textbf{\ref{fig_aver_regret_distinct}}. The average regret illustrates the gap between the average number of nodes activated by the 'optimal' seed set and the selected seed set, where the 'optimal' seed set is the nodes chosen by the oracle algorithm under the true influence probabilities, while the selected seed set is the nodes chosen under the estimated influence probabilities. In the experiment, the maximal number of selected seed nodes $K$ at each round is set as 10. The \textit{IMLinUCB} algorithm and the \textit{Ensemble-Rand-LinThompson} algorithm are both contextual based algorithm with linear assumption on the influence probability, and the contextual information of both algorithms are constructed using the Laplacian approach. 

As could be viewed from \textbf{Figure}  \textbf{\ref{fig_aver_inf_distinct}} and \textbf{Figure}  \textbf{\ref{fig_aver_regret_distinct}},  \textit{Ensemble-Rand-LinThompson} algorithm outperforms the state-of-the-art algorithms including the contextual dependent \textit{IMLinUCB} algorithm, but not significantly. The \textit{Ensemble-Rand-Mean} algorithm has a competitive performance compared with the state-of-the-art algorithms on various datasets. Specifically, the performance of the simple algorithm \textit{Ensemble-Rand-Mean} is significantly better than two contextual based algorithms \textit{IMLinUCB} and \textit{Ensemble-Rand-LinThompson}, without the assumption about the linear form of the influence probability and does not use the contextual information. Commonly, the introduction of contextual information could assist the learner in learning the underlying function and thereby improve the performance of the algorithm. The performance improvement over the contextual based algorithm illustrates the strength of the \textit{Ensemble-Rand-LinThompson} algorithm.

  \begin{figure*} 
\centering 
 \subfigure[Average Regret]{
  \label{regret_mean_plus} 
 \includegraphics[width = 0.41\linewidth]{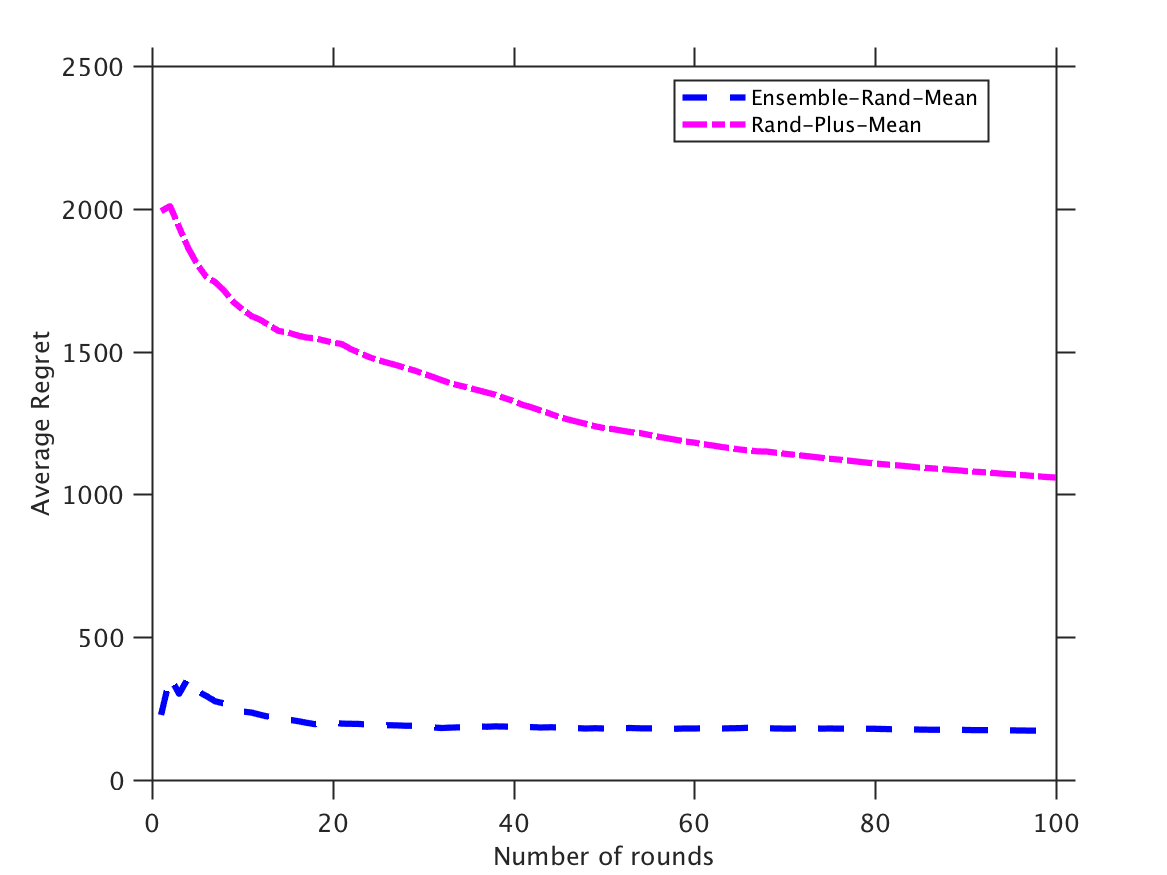}} %
\subfigure[Average Influence Spread]{
 \label{reward_mean_plus} 
 \includegraphics[width = 0.41\linewidth]{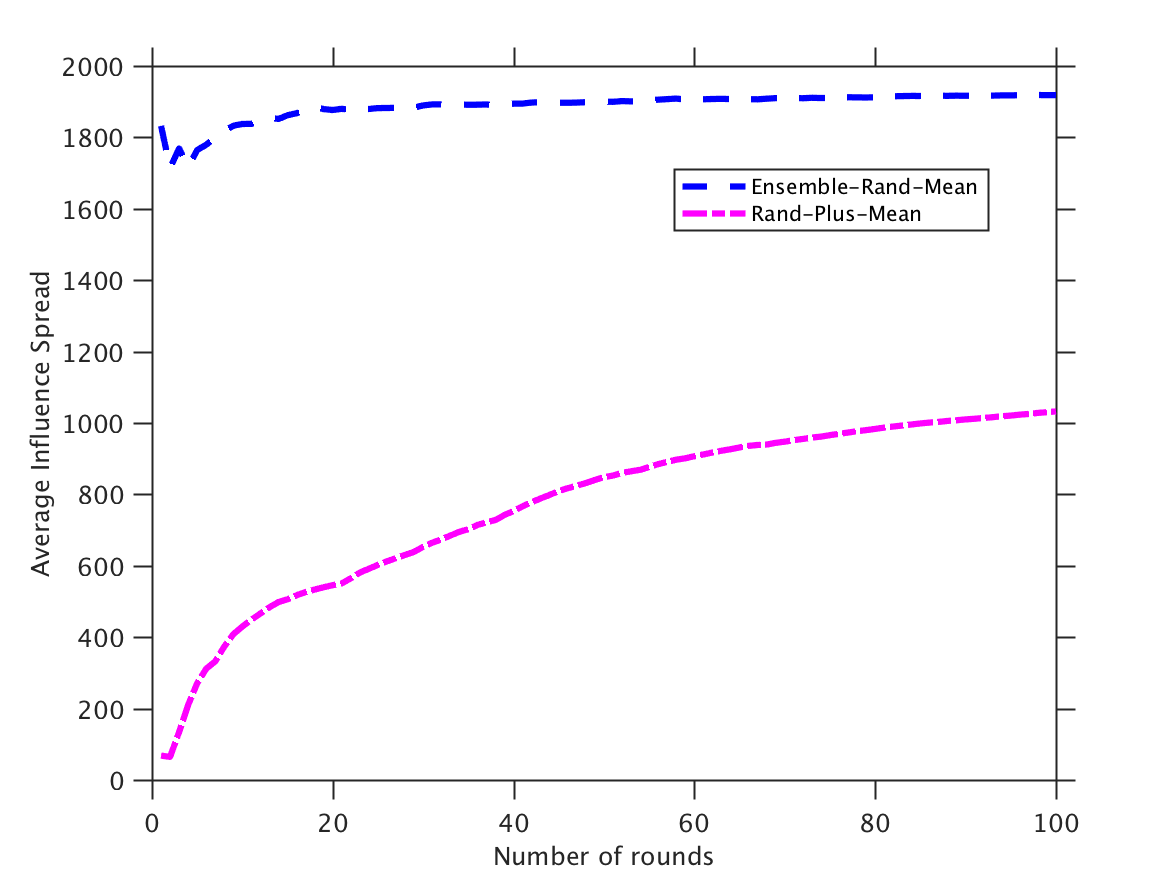}} 
 \caption{\textit{Ensemble-Rand-Mean} V.S. \textit{Explore-Rand-Plus} on Facebook dataset}
 \label{figRandMean_RandPlus} 
 \end{figure*}

Viewing the performance of the proposed algorithms, one may easily raise the following questions,

\begin{itemize}
\item  Does the pure exploit part or the pure explore part perform better than the Rand-Mean algorithm?
\item Is the Ensemble Learning strategy effective for improving the performance the algorithm? Or what is the advantage of the Ensemble Learning algorithm?
\item The \textit{Ensemble-Rand-Mean} algorithm is motivated from the Epsilon-Greedy algorithm, so is Epsilon-Greedy algorithm also very powerful for solving the OIM problem?
\end{itemize}

We conduct the following experiments to answer the above questions.

\begin{figure*} 
\centering 
\subfigure[Facebook]{
 \label{subf_averegret} 
 \includegraphics[width = 0.41\linewidth]{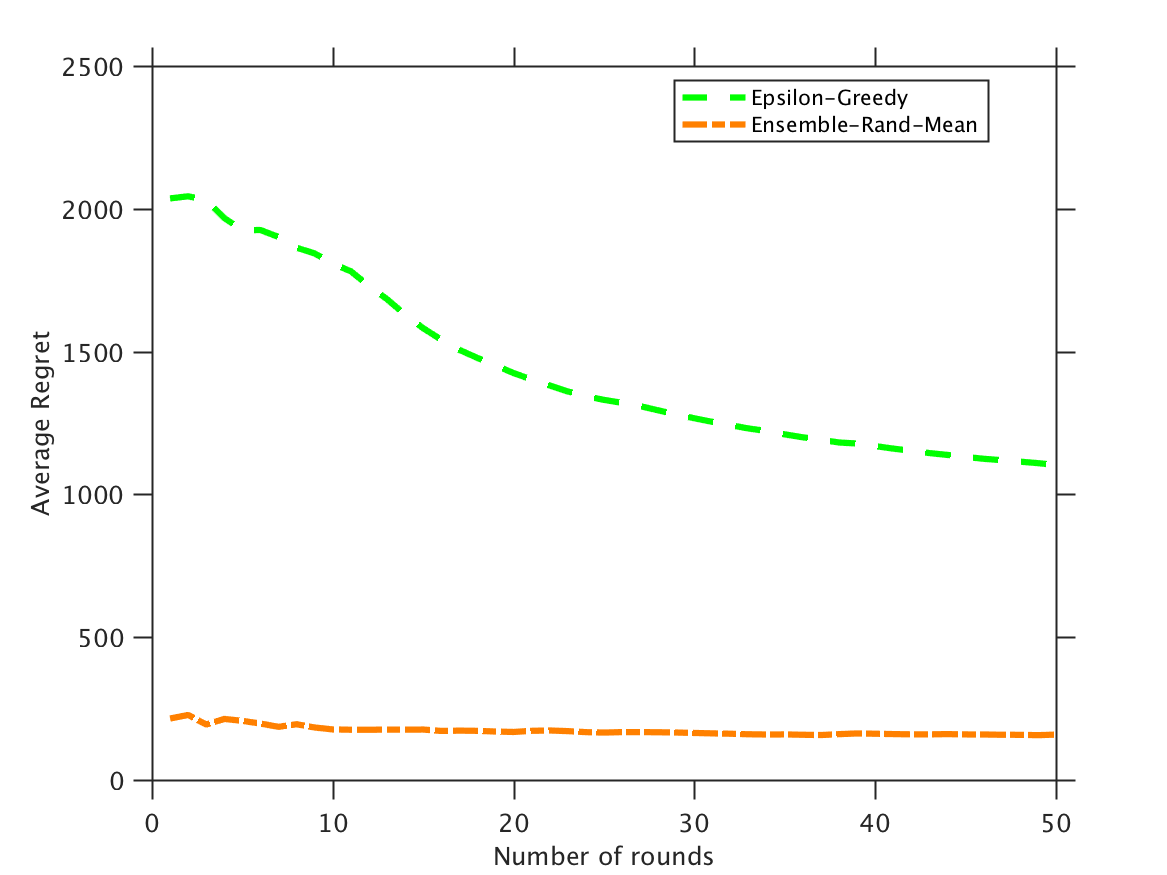}} 
 \subfigure[NetHEPT]{
  \label{subf_avereward} 
 \includegraphics[width = 0.41\linewidth]{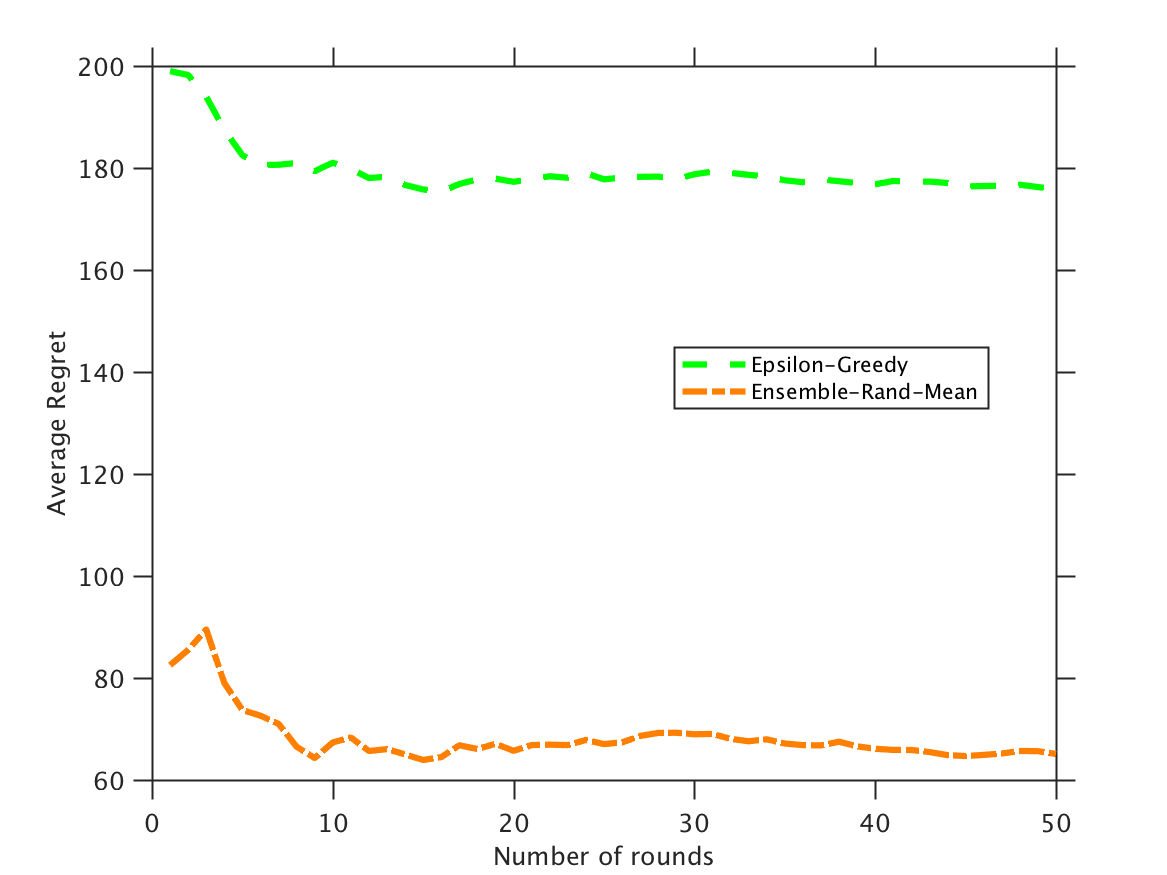}} 
 \caption{The performance of \textit{Epsilon-Greedy} algorithm V.S. \textit{Ensemble-Rand-Mean} algorithm}
 \label{fig_epgreedy_randmean} 
 \end{figure*}

\subsection{The Effect of Automatic Ensemble Learning}

To illustrate the power of the automatic ensemble strategy, we conduct another experiment, which compares the performance of the proposed ensemble algorithms with the corresponding components.

The \textit{Ensemble-Rand-Mean} algorithm is composed of the exploration operation and the exploitation operation. We firstly compare the performance of the \textit{Ensemble-Rand-Mean} algorithm with the algorithm that purely explore with randomly generated influence probabilities, and the algorithm that purely exploit the empirical mean as the estimation of the influence probabilities, which are referred to as the \textit{Explore-Rand} algorithm and the \textit{Exploit-Mean} algorithm separately.   The experiment results are displayed in \textbf{Figure} \textbf{\ref{figRandMean_Rand_Mean}}. 

For the Facebook dataset, the \textit{Explore-Rand} has a better performance than the \textit{Exploit-Mean} algorithm, and the \textit{Ensemble-Rand-Mean} algorithm performs better than both the \textit{Explore-Rand} algorithm and the \textit{Exploit-Mean} algorithm. While for a smaller subgraph of the Facebook dataset, the \textit{Exploit-Mean} algorithm has a poor performance at the earlier rounds since the influence function has not been learned very well, while in later rounds it performs better than the \textit{Explore-Rand} algorithm due to the accumulated experience with the increasing number of rounds. The \textit{Ensemble-Rand-Mean} algorithm automatically makes up the shortage of both algorithms, and adapts well with distinct datasets, showing the robustness of the \textit{Ensemble-Rand-Mean} algorithm.

Note that for the \textit{Ensemble-Rand-Mean} algorithm, only one strategy could be exploited at each round. We further compare the \textit{Ensemble-Rand-Mean} algorithm with the \textit{Rand-Plus-Mean} algorithm, which estimates the influence probability using the summation of the empirical mean and the randomly generated value. As can be viewed in \textbf{Figure} \textbf{\ref{figRandMean_RandPlus}}, the  \textit{Ensemble-Rand-Mean} algorithm obtains significant improvement over the \textit{Rand-Plus-Mean} in terms of both the average regret and the average influence spread, illustrating the effectiveness of the automatic ensemble strategy.

\begin{figure} 
\centering 
 \label{subf_averegret} 
 \includegraphics[width = 0.95\linewidth]{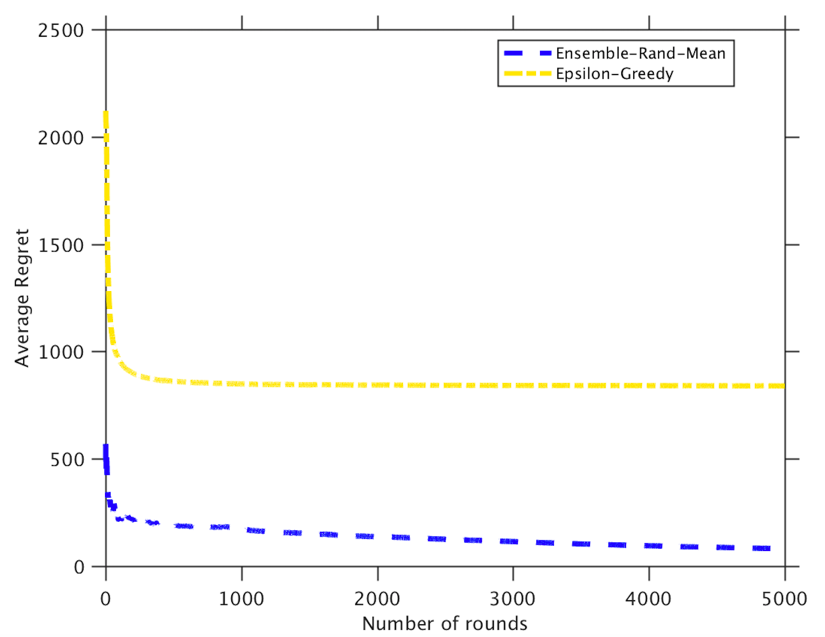} 
 \caption{The performance of \textit{Ensemble-Rand-Mean} compared with \textit{Epsilon-Greedy}}
 \label{figure_epsilongreedy_moreRounds} 
 \end{figure}

\begin{figure*} 
\centering 
\subfigure[Average Regret]{
 \label{subf_averegret} 
 \includegraphics[width = 0.41\linewidth]{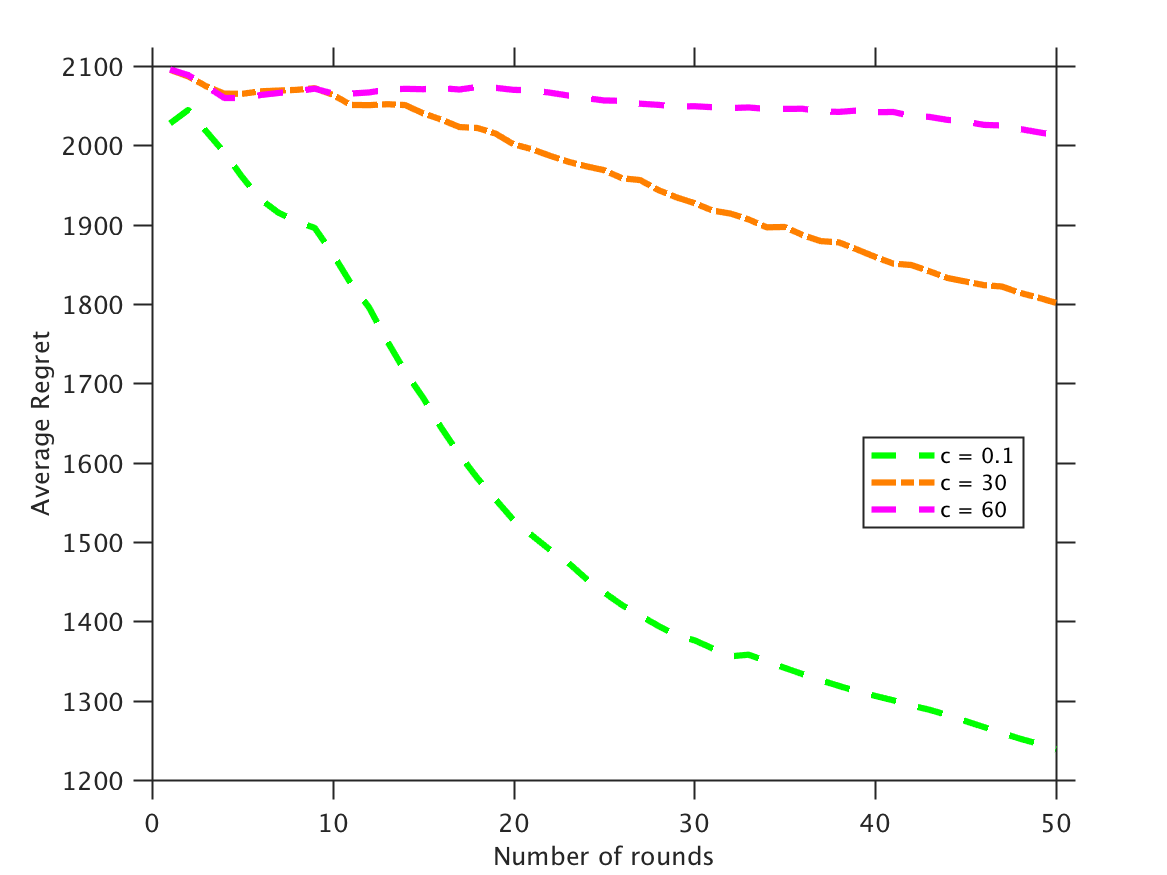}} 
 \subfigure[Average Influence Spread]{
  \label{subf_avereward} 
 \includegraphics[width = 0.41\linewidth]{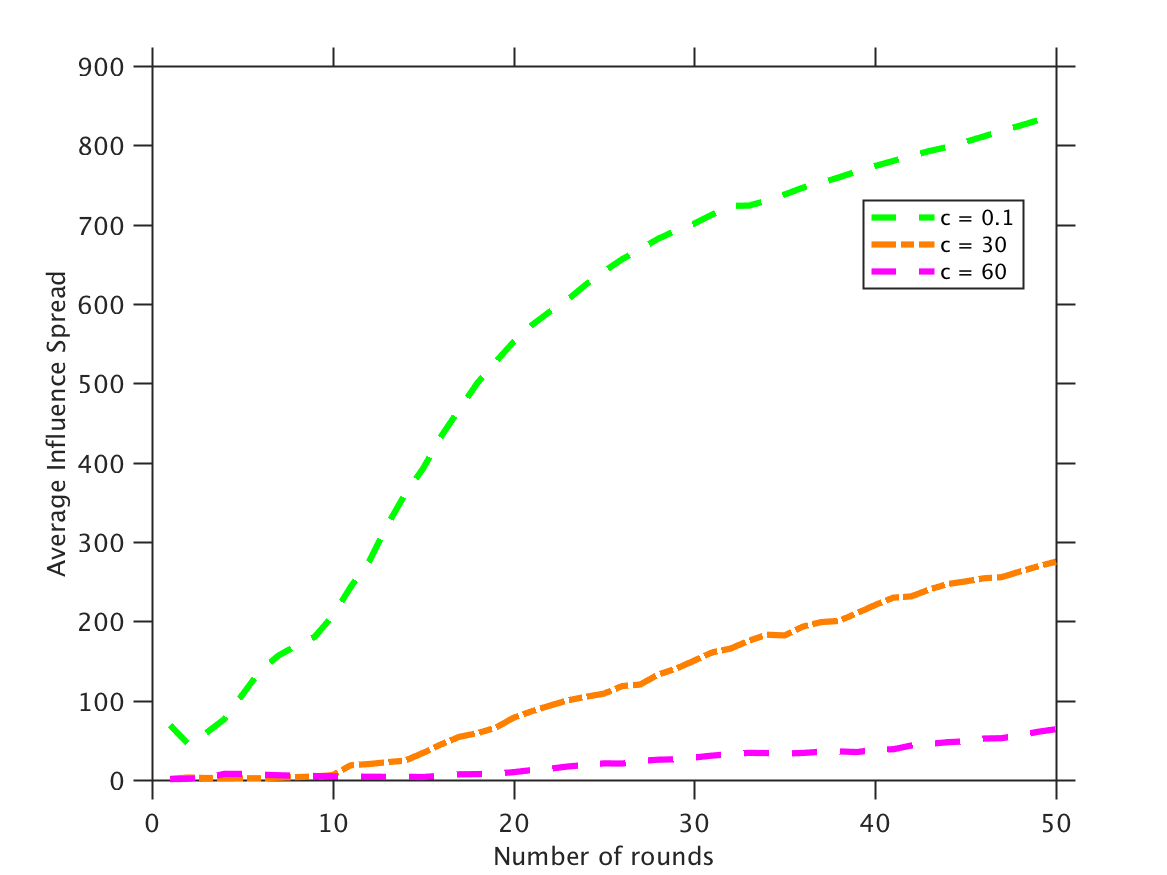}} 
 \caption{The performance of \textit{Epsilon-Greedy} algorithm under different parameters on Facebook dataset}
 \label{figure_epsilongreedy} 
 \end{figure*}

\subsection{Performance of Epsilon-Greedy}
The first experiment shows that the \textit{Ensemble-Rand-Mean} algorithm achieves significant improvement over the state-of-the-art algorithms. Note that the invention of \textit{Ensemble-Rand-Mean} algorithm is motivated from the \textit{Epsilon-Greedy} algorithm, which selects the best strategy by exploiting the empirical mean with probability 1-$\epsilon_{t}$, and explores uniformly over all other actions with probability $\epsilon_{t}$ at round $t$. One might wonder is the performance of this algorithm, is this algorithm also powerful for solving the OIM problem? In the third experiment, we show the performance of the \textit{Epsilon-Greedy} algorithm. It was shown by \cite{chen2013combinatorial} that logarithmic regret could be guaranteed if $\epsilon_{t}$ is set as $c/t$ for the OIM problem under the edge-level feedback, where $c$ is a constant. We tried distinct settings of $c$, and finally select $c = 0.1$ at which \textit{Epsilon-Greedy} performs the best on the Facebook dataset among the selected parameters. As can be viewed from \textbf{Figure} \textbf{\ref{fig_epgreedy_randmean}}, the \textit{Epsilon-Greedy} algorithm has a poor performance compared with the \textit{Ensemble-Rand-Mean} algorithm on both Facebook dataset and the larger NetHEPT dataset. The comparison of these two algorithms with more rounds is shown in \textbf{Figure} \textbf{\ref{figure_epsilongreedy_moreRounds}}.

Besides, even though it has been demonstrated theoretically that the algorithm could have a good performance setting $\epsilon_{t}$ as $c/t$, it still requires to select a suitable parameter for distinct datasets, which is rather hard and requires lots of energy. To illustrate the effect of distinct settings of the parameter on the performance of the \textit{Epsilon-Greedy} algorithm, we compare the cases when the parameter $c$ is $15$, $25$, and $35$. From the experiment results shown in \textbf{Figure} \textbf{\ref{figure_epsilongreedy}}, we can observe that when the parameter $c$ is set as $0.1$, the average regret of the \textit{Epsilon-Greedy} algorithm is smaller than the other two cases and converges in a faster speed.

\section{Conclusion}
In this work, we propose a novel algorithm for the influence maximization problem, and show that the performance of our algorithm is significantly better than the state-of-the-art algorithms. It is worth noting that previous works mainly assume that the selected influentials have identical payoffs. But in reality, distinct advertisers might have different wages due to their various experiences and abilities, which is a considerable direction for future research. In addition, the newly proposed methods including decision trees could also be utilized for the online influence maximization problem. Considering that gaussian process regression merges both the rule-based and similarity-based theories, using the gaussian process regression approach to solve the online influence maximization problem could be another future direction. Besides, it is sometimes hard to obtain the whole topology of the social network in practical. However, it is easier to obtain the closest friends of a given person, which corresponds to the nearby connections in a social network. Thus it is meaningful to investigate the influence maximization problem under this kind of incomplete graph.
 
\section{Acknowledgements}
Xiaojin Zhang would like to thank Shengyu Zhang for helpful discussions.

\clearpage

\bibliographystyle{plain}
\bibliography{referencewww} 

\begin{thebibliography}{10}

\bibitem{abbasi2011improved}
Yasin Abbasi-Yadkori, D{\'a}vid P{\'a}l, and Csaba Szepesv{\'a}ri.
\newblock Improved algorithms for linear stochastic bandits.
\newblock In {\em Advances in Neural Information Processing Systems}, pages
  2312--2320, 2011.

\bibitem{abeille2017linear}
Marc Abeille and Alessandro Lazaric.
\newblock Linear thompson sampling revisited.
\newblock In {\em AISTATS 2017-20th International Conference on Artificial
  Intelligence and Statistics}, 2017.

\bibitem{agrawal2012analysis}
Shipra Agrawal and Navin Goyal.
\newblock Analysis of thompson sampling for the multi-armed bandit problem.
\newblock In {\em Conference on Learning Theory}, pages 39--1, 2012.

\bibitem{agrawal2013thompson}
Shipra Agrawal and Navin Goyal.
\newblock Thompson sampling for contextual bandits with linear payoffs.
\newblock In {\em International Conference on Machine Learning}, pages
  127--135, 2013.

\bibitem{carpentier2016revealing}
Alexandra Carpentier and Michal Valko.
\newblock Revealing graph bandits for maximizing local influence.
\newblock In {\em Artificial Intelligence and Statistics}, pages 10--18, 2016.

\bibitem{chapelle2011empirical}
Olivier Chapelle and Lihong Li.
\newblock An empirical evaluation of thompson sampling.
\newblock In {\em Advances in neural information processing systems}, pages
  2249--2257, 2011.

\bibitem{chen2010scalable}
Wei Chen, Chi Wang, and Yajun Wang.
\newblock Scalable influence maximization for prevalent viral marketing in
  large-scale social networks.
\newblock In {\em Proceedings of the 16th ACM SIGKDD international conference
  on Knowledge discovery and data mining}, pages 1029--1038. ACM, 2010.

\bibitem{chen2009efficient}
Wei Chen, Yajun Wang, and Siyu Yang.
\newblock Efficient influence maximization in social networks.
\newblock In {\em Proceedings of the 15th ACM SIGKDD international conference
  on Knowledge discovery and data mining}, pages 199--208. ACM, 2009.

\bibitem{chen2013combinatorial}
Wei Chen, Yajun Wang, and Yang Yuan.
\newblock Combinatorial multi-armed bandit: General framework and applications.
\newblock In {\em International Conference on Machine Learning}, pages
  151--159, 2013.

\bibitem{chen2016combinatorial}
Wei Chen, Yajun Wang, Yang Yuan, and Qinshi Wang.
\newblock Combinatorial multi-armed bandit and its extension to
  probabilistically triggered arms.
\newblock {\em The Journal of Machine Learning Research}, 17(1):1746--1778,
  2016.

\bibitem{domingos2001mining}
Pedro Domingos and Matt Richardson.
\newblock Mining the network value of customers.
\newblock In {\em Proceedings of the seventh ACM SIGKDD international
  conference on Knowledge discovery and data mining}, pages 57--66. ACM, 2001.

\bibitem{fang2014networked}
Meng Fang and Dacheng Tao.
\newblock Networked bandits with disjoint linear payoffs.
\newblock In {\em Proceedings of the 20th ACM SIGKDD international conference
  on Knowledge discovery and data mining}, pages 1106--1115. ACM, 2014.

\bibitem{goyal2011data}
Amit Goyal, Francesco Bonchi, and Laks~VS Lakshmanan.
\newblock A data-based approach to social influence maximization.
\newblock {\em Proceedings of the VLDB Endowment}, 5(1):73--84, 2011.

\bibitem{goyal2011celf++}
Amit Goyal, Wei Lu, and Laks~VS Lakshmanan.
\newblock Celf++: optimizing the greedy algorithm for influence maximization in
  social networks.
\newblock In {\em Proceedings of the 20th international conference companion on
  World wide web}, pages 47--48. ACM, 2011.

\bibitem{jung2012irie}
Kyomin Jung, Wooram Heo, and Wei Chen.
\newblock Irie: Scalable and robust influence maximization in social networks.
\newblock In {\em Data Mining (ICDM), 2012 IEEE 12th International Conference
  on}, pages 918--923. IEEE, 2012.

\bibitem{kaufmann2012thompson}
Emilie Kaufmann, Nathaniel Korda, and R{\'e}mi Munos.
\newblock Thompson sampling: An asymptotically optimal finite-time analysis.
\newblock In {\em ALT}, volume~12, pages 199--213. Springer, 2012.

\bibitem{kempe2003maximizing}
David Kempe, Jon Kleinberg, and {\'E}va Tardos.
\newblock Maximizing the spread of influence through a social network.
\newblock In {\em Proceedings of the ninth ACM SIGKDD international conference
  on Knowledge discovery and data mining}, pages 137--146. ACM, 2003.

\bibitem{kim2013scalable}
Jinha Kim, Seung-Keol Kim, and Hwanjo Yu.
\newblock Scalable and parallelizable processing of influence maximization for
  large-scale social networks?
\newblock In {\em Data Engineering (ICDE), 2013 IEEE 29th International
  Conference on}, pages 266--277. IEEE, 2013.

\bibitem{korda2013thompson}
Nathaniel Korda, Emilie Kaufmann, and Remi Munos.
\newblock Thompson sampling for 1-dimensional exponential family bandits.
\newblock In {\em Advances in Neural Information Processing Systems}, pages
  1448--1456, 2013.

\bibitem{lei2015online}
Siyu Lei, Silviu Maniu, Luyi Mo, Reynold Cheng, and Pierre Senellart.
\newblock Online influence maximization.
\newblock In {\em Proceedings of the 21th ACM SIGKDD International Conference
  on Knowledge Discovery and Data Mining}, pages 645--654. ACM, 2015.

\bibitem{leskovec2007cost}
Jure Leskovec, Andreas Krause, Carlos Guestrin, Christos Faloutsos, Jeanne
  VanBriesen, and Natalie Glance.
\newblock Cost-effective outbreak detection in networks.
\newblock In {\em Proceedings of the 13th ACM SIGKDD international conference
  on Knowledge discovery and data mining}, pages 420--429. ACM, 2007.

\bibitem{richardson2002mining}
Matthew Richardson and Pedro Domingos.
\newblock Mining knowledge-sharing sites for viral marketing.
\newblock In {\em Proceedings of the eighth ACM SIGKDD international conference
  on Knowledge discovery and data mining}, pages 61--70. ACM, 2002.

\bibitem{russo2016information}
Daniel Russo and Benjamin Van~Roy.
\newblock An information-theoretic analysis of thompson sampling.
\newblock {\em The Journal of Machine Learning Research}, 17(1):2442--2471,
  2016.

\bibitem{saritacc2016online}
{\"O}mer Sar{\i}ta{\c{c}}, Altu{\u{g}} Karakurt, and Cem Tekin.
\newblock Online contextual influence maximization in social networks.
\newblock In {\em Communication, Control, and Computing (Allerton), 2016 54th
  Annual Allerton Conference on}, pages 1204--1211. IEEE, 2016.

\bibitem{scott2010modern}
Steven~L Scott.
\newblock A modern bayesian look at the multi-armed bandit.
\newblock {\em Applied Stochastic Models in Business and Industry},
  26(6):639--658, 2010.

\bibitem{seldin2012pac}
Yevgeny Seldin, Nicol{\`o} Cesa-Bianchi, Peter Auer, Fran{\c{c}}ois Laviolette,
  and John Shawe-Taylor.
\newblock Pac-bayes-bernstein inequality for martingales and its application to
  multiarmed bandits.
\newblock In {\em Proceedings of the Workshop on On-line Trading of Exploration
  and Exploitation 2}, pages 98--111, 2012.

\bibitem{tang2015influence}
Youze Tang, Yanchen Shi, and Xiaokui Xiao.
\newblock Influence maximization in near-linear time: A martingale approach.
\newblock In {\em Proceedings of the 2015 ACM SIGMOD International Conference
  on Management of Data}, pages 1539--1554. ACM, 2015.

\bibitem{tang2014influence}
Youze Tang, Xiaokui Xiao, and Yanchen Shi.
\newblock Influence maximization: Near-optimal time complexity meets practical
  efficiency.
\newblock In {\em Proceedings of the 2014 ACM SIGMOD international conference
  on Management of data}, pages 75--86. ACM, 2014.

\bibitem{tossou2017thompson}
Aristide~CY Tossou, Christos Dimitrakakis, and Devdatt~P Dubhashi.
\newblock Thompson sampling for stochastic bandits with graph feedback.
\newblock In {\em AAAI}, pages 2660--2666, 2017.

\bibitem{valko2016bandits}
Michal Valko.
\newblock {\em Bandits on graphs and structures}.
\newblock PhD thesis, {\'E}cole normale sup{\'e}rieure de Cachan-ENS Cachan,
  2016.

\bibitem{vaswani2017diffusion}
Sharan Vaswani, Branislav Kveton, Zheng Wen, Mohammad Ghavamzadeh, Laks
  Lakshmanan, and Mark Schmidt.
\newblock Diffusion independent semi-bandit influence maximization.
\newblock {\em arXiv preprint arXiv:1703.00557}, 2017.

\bibitem{vaswani2015influence}
Sharan Vaswani, Laks Lakshmanan, Mark Schmidt, et~al.
\newblock Influence maximization with bandits.
\newblock {\em arXiv preprint arXiv:1503.00024}, 2015.

\bibitem{wen2016influence}
Zheng Wen, Branislav Kveton, and Michal Valko.
\newblock Influence maximization with semi-bandit feedback.
\newblock {\em arXiv preprint arXiv:1605.06593}, 2016.

\end{thebibliography}

\end{document}